\RequirePackage{fix-cm}
\documentclass[twocolumn, natbib]{svjour3}          
\smartqed  
\usepackage[misc]{ifsym}
\usepackage{color}
\usepackage[table,dvipsnames]{xcolor}
\usepackage{tikz}  %
\usepackage{times}
\usepackage{graphicx}
\usepackage{amsmath}
\usepackage{amssymb}
\usepackage{multirow}
\usepackage{textcomp}
\usepackage{natbib}
\usepackage{url}
\usepackage{comment}
\renewcommand{\cite}{\citep}
\usepackage[colorlinks,
            linkcolor=red,
            anchorcolor=blue,
            citecolor=blue]{hyperref}

\usepackage{subfiles}
\usepackage[normalem]{ulem}
\usepackage{pgfplots} %
\usepackage{pgfplotstable} %
\pgfplotsset{compat=newest} %
\usetikzlibrary{plotmarks} %
\usepackage[group-separator={,},output-decimal-marker = {.}, group-minimum-digits={3}]{siunitx} %
\usepackage{subfigure}
\usepackage{paralist}
\usepackage{ragged2e}
\usepackage{appendix}
\include{packages}
\include{macros}
\journalname{}
\begin{document}

\title{Polysemy Deciphering Network for Robust Human-Object Interaction Detection
}


\author{Xubin Zhong\textsuperscript{1}       \and
        Changxing Ding\textsuperscript{1,2} 
         \and Xian Qu\textsuperscript{1}
          \and Dacheng Tao\textsuperscript{3}
}


\institute{\Letter \ Changxing Ding\textsuperscript{1,2}  \at
          \email{chxding@scut.edu.cn}
          \and
          Xubin Zhong\textsuperscript{1}  \at
          \email{eexubin@mail.scut.edu.cn}
          \and
          Xian Qu\textsuperscript{1}  \at
          \email{eequxian.scut@mail.scut.edu.cn}
          \and
          Dacheng Tao\textsuperscript{3}  \at
          \email{dacheng.tao@jd.com}
          \and
           $^{1}$ \ School of Electronic and Information Engineering, South China University of Technology, Guangzhou 510000, China \\
           $^{2}$ Pazhou Lab, Guangzhou 510330, China \\
           $^{3}$  JD Explore Academy at JD.com, Beijing, China
}

\date{Received: date / Accepted: date}

\maketitle

\begin{abstract}
Human-Object Interaction (HOI) detection is important to human-centric scene understanding tasks. Existing works tend to assume that the same verb  has similar visual characteristics in different HOI categories, an approach that ignores the diverse semantic meanings of the verb. To address this issue, in this paper, we propose a novel Polysemy Deciphering Network (PD-Net) that decodes the visual polysemy of verbs for HOI detection in three distinct ways. First, we refine features for HOI detection to be polysemy-aware through the use of two novel modules: namely, Language Prior-guided Channel Attention (LPCA) and Language Prior-based Feature Augmentation (LPFA). LPCA highlights important elements in human and object appearance features for each HOI category to be identified; moreover, LPFA augments human pose and spatial features for HOI detection using language priors, enabling the verb classifiers to receive language hints that reduce intra-class variation for the same verb. Second, we introduce a novel Polysemy-Aware Modal Fusion module (PAMF), which guides PD-Net to make decisions based on feature types deemed  more important according to the language priors. Third, we propose to relieve the verb polysemy problem through sharing verb classifiers for semantically similar HOI categories. Furthermore, to expedite research on the verb polysemy problem, we build a new benchmark dataset named HOI-VerbPolysemy (HOI-VP), which includes common verbs (predicates) that have diverse semantic meanings in the real world. Finally, through deciphering the visual polysemy of verbs, our approach is demonstrated to outperform state-of-the-art methods by significant margins on the HICO-DET, V-COCO, and HOI-VP databases.
Code and data in this paper are available at  \url{https://github.com/MuchHair/PD-Net}.
\end{abstract}

\section{Introduction}
\label{sec:intro}
\begin{figure}[t]
     \centering
\subfigure[person play  soccer\_ball]{
    \includegraphics[width=0.22\textwidth]{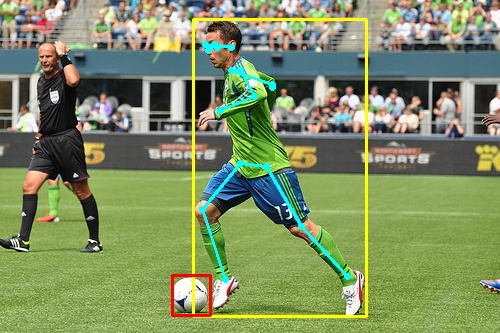}
}
     \subfigure[person play frisbee]{
    \includegraphics[width=0.22\textwidth]{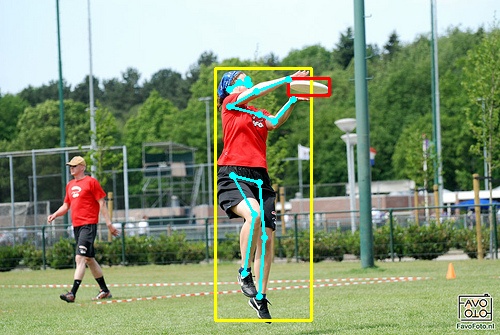}
}

 \subfigure[person hold book]{
    \includegraphics[width=0.22\textwidth]{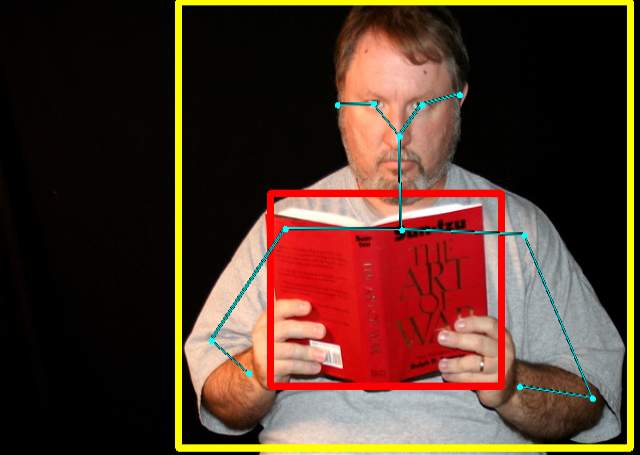}
}
\subfigure[person hold elephant]{
  \includegraphics[width=0.22\textwidth]{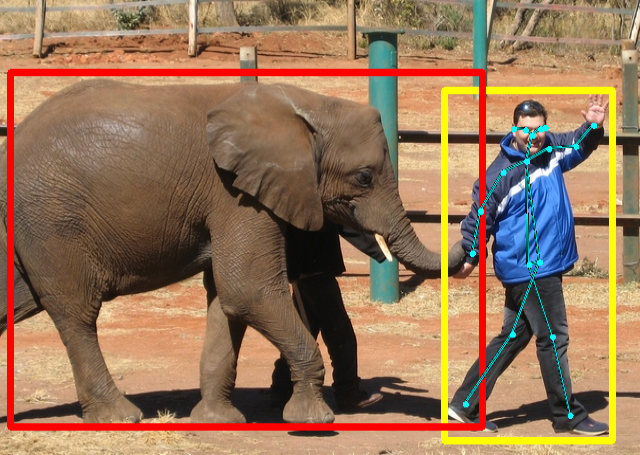}
}

\subfigure[person fly kite]{
    \includegraphics[width=0.22\textwidth]{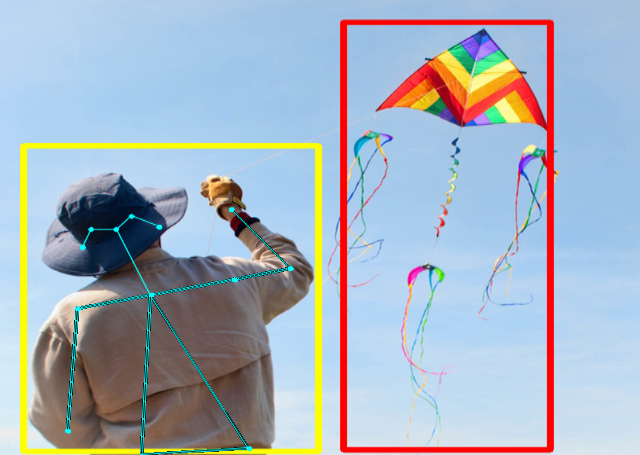}
}
    \subfigure[person fly airplane]{
    \includegraphics[width=0.22\textwidth]{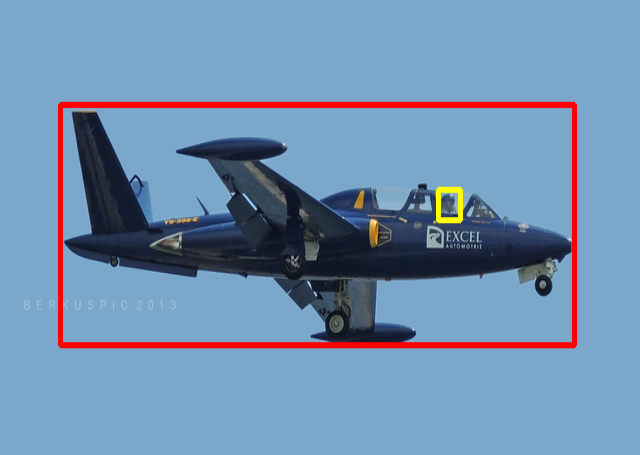}
}

   \caption{Examples reflecting the verb polysemy problem in HOI detection. In terms of describing the HOIs, (a) and (b) present HOI examples of ``play''. ``feet'' are more important in (a) while  ``hands'' are more important in (b). (c) and (d) illustrate HOI examples of ``hold''. The human-object pairs in (c) and (d) are characterized by dramatically different human-object spatial features, i.e. the relative location between two bounding boxes. (e) and (f) illustrate HOI examples of ``fly''. The ``person'' in (e) exhibits discriminative pose features while the ``person'' in (f) does not. }
    \label{Figure:diverse_semantic_role}
\end{figure}

In recent years, researchers working in the field of computer vision have begun to pay increasing attention to scene understanding tasks \cite{ISI:000351518500007,lu2016visual,zhang2017visual,ISI:000511490100004, lin2020gps}. Since human beings are often central to real-world scenes, Human-Object Interaction (HOI) detection has become a fundamental problem in scene understanding. HOI detection involves not only identifying the classes and locations of objects in the images, but also the interactions (verbs) between each human-object pair. As shown in Fig. \ref{Figure:diverse_semantic_role}, an interaction between a human-object pair can be represented by a triplet $<$$person$ $verb$ $object$$>$, herein referred to as one HOI category. One human-object pair may comprise multiple triplets, e.g. $<$$person$ $fly$ $airplane$$>$ and $<$$person$ $ride$ $airplane$$>$.

The HOI detection task is notably challenging \cite{chao2018learning, Gao2018iCANIA}. One major reason is that verbs can be polysemic. As illustrated in Fig. \ref{Figure:diverse_semantic_role}, a verb may convey substantially different semantic meanings and visual characteristics with respect to different objects, as these objects may have diverse functions and attributes. One pair of examples can be found in Fig. \ref{Figure:diverse_semantic_role}(a) and (b). Here, the  ``feet''  are the more discriminative parts of the human figure for $<$$person$ $play$ $soccer\_ball$$>$ while ``hands'' are more important for describing $<$$person$ $play$ $frisbee$$>$. A second pair of examples is presented in Fig. \ref{Figure:diverse_semantic_role}(c) and (d). The human-object pairs in (c) and (d), despite being tagged with the same verb, present dramatically different human-object spatial features. Another more serious consideration is that the importance of the same type of visual feature may vary dramatically as the objects of interest change. For example, the human pose plays a vital role in describing $<$$person$ $fly$ $kite$$>$ in Fig. \ref{Figure:diverse_semantic_role}(e); by contrast, the human pose is invisible and therefore useless for characterizing $<$$person$ $fly$ $airplane$$>$ in Fig. \ref{Figure:diverse_semantic_role}(f).  Verb polysemy therefore presents a significant challenge in the HOI detection.

\justifying
The problem of verb polysemy is relatively underexplored, and sometimes even ignored, in existing works \cite{xu2019learning, li2019transferable, liao2020ppdm, wang2020learning}. Most contemporary approaches tend to assume that the same verb will have similar visual characteristics across different HOI categories, and accordingly opt to design object-shared verb classifiers. When the verb classifier is shared among all objects, each verb obtains more training samples, thereby promoting the robustness of the classification for HOI categories with a small sample size. However, due to the polysemic nature of the verbs, a dramatic semantic gap may exist between instances of the same verb across different HOI categories. Chao \MakeLowercase{\textit{et al.}} \citeyearpar{chao2018learning} constructed object-specific verb classifiers for each HOI category, which are able to overcome the polysemy problem for HOI categories that have sufficient training samples. However, this approach lacks few- and zero-shot learning abilities for HOI categories where only small amounts of training data are available.

\begin{figure*}[t]
\centering
\includegraphics[width=0.8\textwidth]{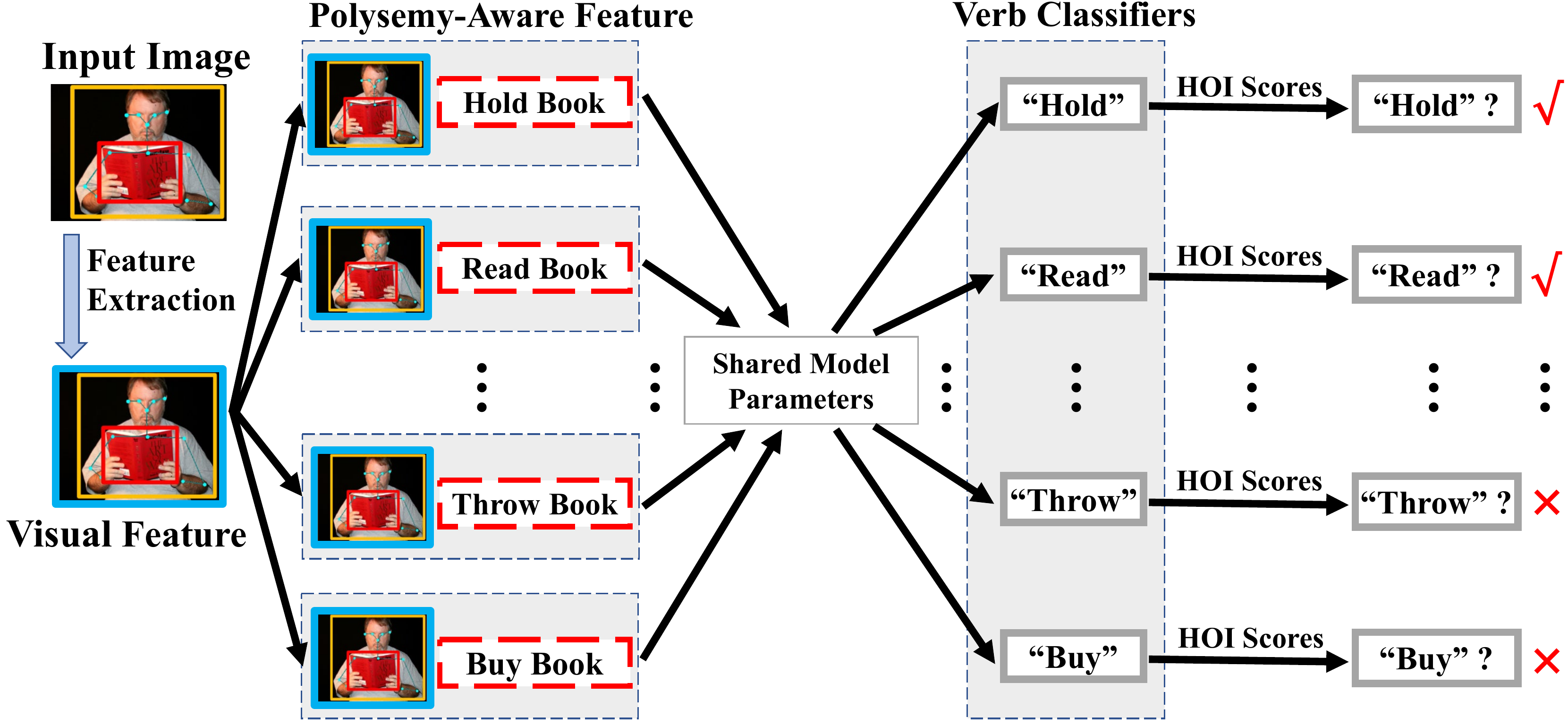}
\caption{
Visual features of each human-object pair are duplicated multiple times so that polysemy-aware visual features can be obtained under the guidance of language priors. Each polysemy-aware feature is sent to a specific verb classifier, which can be any type of the three verb classifiers mentioned in this paper. To reduce the number of duplicated human-object pairs, meaningless HOI categories (e.g. $<$$person$ $eat$ $book$$>$ and $<$$person$ $ride$ $book$$>$) are ignored. Meaningful and common HOI categories (e.g. $<$$person$ $hold$ $book$$>$ and $<$$person$ $read$ $book$$>$) are available in each popular HOI detection database.  Multiple verbs are ``checked'' in this figure as HOI detection performs multi-label verb classification for each human-object pair.}
\label{Figure:singlebinary}
\end{figure*}

In this paper, we propose a novel Polysemy Deciphering Network (PD-Net) to address the challenging verb polysemy problem. As illustrated in Fig. \ref{Figure:singlebinary}, PD-Net transforms the multi-label verb classifications for each human-object pair into a set of binary classification problems. Here, each binary classifier is used for the verification of one verb category. The classifiers share the majority of their parameters; the main difference lies in the input features. Next, we decode the verb polysemy in the following three ways.

First, we enable features sent for each binary classifier to be polysemy-aware using two novel modules, namely Language Prior-guided Channel Attention (LPCA) and Language Prior-based Feature Augmentation (LPFA). The language priors are two word embeddings made up  of one verb and one object. The object class is predicted by one object detector; the verb is the one to be determined by one specific binary verb classifier. For its part, LPCA is applied to both the human and object appearance features. The two appearance features are usually redundant, as only part of their information is involved for one specific HOI category (see Fig. \ref{Figure:diverse_semantic_role}). Therefore, LPCA is used to highlight important elements in the appearance features for each binary classifier. Moreover, both human-object spatial and human pose features are often vague and can vary dramatically for the same verb, as shown in Fig. \ref{Figure:diverse_semantic_role}(a) and (b), (c) and (d); we therefore propose LPFA, which concatenates the two features with language priors, respectively. In this way, the classifiers can receive hints to reduce the intra-class variation of the same verb for the pose and spatial features.

We further design a novel Polysemy-Aware Modal Fusion module (PAMF), which produces attention scores based on the above language priors in order to dynamically fuse multiple feature types. The language priors provide hints regarding the importance of the features for each HOI category. As can be seen in Fig. \ref{Figure:diverse_semantic_role}, the human pose feature is discriminative when the language prior is ``fly kite'' (Fig. \ref{Figure:diverse_semantic_role}(e)), but is less useful when the language prior is ``fly airplane'' (Fig. \ref{Figure:diverse_semantic_role}(f)). Therefore, our proposed PAMF deciphers the verb polysemy problem by highlighting the features that are more important for each HOI category.

Moreover, as mentioned above, both  object-shared and object-specific verb classifiers have limitations. We therefore propose a novel clustering-based object-specific verb classifier, which combines the advantages of object-shared and object-specific classifiers. The main motivation is to ensure that semantically similar HOI categories containing the same verb, e.g. $<$$person$ $hold$ $cow$$>$ and $<$$person$ $hold$ $elephant$$>$ can share the same verb classifier. HOIs that are semantically very different (e.g. $<$$person$ $hold$ $book$$>$ and $<$$person$ $hold$ $backpack$$>$) are identified using another verb classifier. In this way, the verb polysemy problem is mitigated. Meanwhile, clustering-based object-specific classifiers has the capacity to handle the few- and zero-shot learning problems that arise in HOI detection, since we merge the training data of semantically similar HOI categories.

To the best of our knowledge, our proposed PD-Net is the first approach to explicitly handle the verb polysemy problem in HOI detection.
More  impressively, our experimental results on three databases demonstrate  that our approach consistently outperforms state-of-the-art methods by considerable margins.
A preliminary version of this paper has been published in~\cite{zhong2020polysemy}.
Compared with the conference version, this version further proposes a novel Language Prior-guided Channel Attention module, simplifies the architecture of PD-Net by using Clustering-based Object-Specific classifiers, builds a new database (named HOI-VP) to facilitate the research on the verb polysemy problem, and includes further experimental investigations.

The remainder of this paper is organized as follows. Section \ref{sec:Related_Works} briefly reviews related works. The details of the proposed components of PD-Net are described in Section \ref{sec:method}. The databases and implementation details are introduced in Section \ref{sec:Experiments_setting}, while the experimental results are presented in Section \ref{sec:EXPERIMENTS}. Finally, we conclude the paper in Section \ref{sec:conclusion}.

\section{Related Works}
\label{sec:Related_Works}
\noindent{\bf Human-Object Interaction Detection.} HOI detection performs multi-label verb classification for each human-object pair, meaning that  the interaction between the same human-object pair may be described using multiple verbs. Depending on the order of verb classification and target object association, existing HOI detection approaches can be divided into two categories. The first category of methods infer the verb actions being performed by one person, then associate each verb with a single object in the image. Multiple target object association approaches have been proposed. For example, Shen \textit{et al.} \citeyearpar{Shen2018Scaling} proposed an approach based on the value of object detection scores, while Gkioxari \textit{et al.} \citeyearpar{Gkioxari2017Detecting} fitted a distribution density function of the target object locations based on the human appearance feature. Moreover, Qi \textit{et al.} \citeyearpar{Qi2018Learning} adopted a graph parsing network to associate the target objects.  Liao \textit{et al.}  \citeyearpar{liao2020ppdm}  and  Wang \textit{et al.} \citeyearpar{wang2020learning} first defined interaction points for HOI detection; next, they locate the interaction points and associate each  point with one human-object pair.

The second category of methods first pair each human instance with all object instances as candidate human-object pairs, then recognize the verb for each candidate pair \cite{gupta2019no}. Many types of features have been employed to promote the verb classification performance.
For example, Wan \textit{et al.} \citeyearpar{wan2019pose} employed both human parts and pose-aware features for verb classification, while Xu \textit{et al.} \citeyearpar{8848601} exploited human gaze and intention  to assist HOI detection.  Furthermore, Wang \textit{et al.} \citeyearpar{wang2019deep} extracted context-aware human and object appearance features to promote HOI detection performance.
Li \textit{et al.}  \citeyearpar{li2020detailed} utilized 3D pose models and 3D object location to assist HOI detection. Moreover, Li \textit{et al.} \citeyearpar{li2020pastanet} annotated large amounts of part-level human-object interactions and trained a PaStaNet,
which is helpful for HOI detection models to make use of fine-grained human part features.
A large number of novel model architectures for HOI detection have been also developed.
For example, Li \textit{et al.} \citeyearpar{li2019transferable} introduced a Transferable Interactiveness Network that suppresses candidate pairs without interactions.
Peyre \MakeLowercase{\textit{et al.}} \citeyearpar{peyre2019detecting} constructed a multi-stream model that projects visual features and word embeddings to a joint space,
which is helpful for unseen HOI category detection.
Xu \textit{et al.} \citeyearpar{xu2019learning} constructed a graph neural network to promote the quality of word embeddings by utilizing the correlation between semantically similar verbs, while Zhou \textit{et al.}  \citeyearpar{zhou2020cascaded} proposed a cascade architecture that facilitates coarse-to-fine HOI detection.

Besides, HOI recognition \cite{chao2015hico} is one similar task to HOI detection. Briefly, HOI recognition methods predict all possible HOI categories in an image but they do not detect the location of involved human-object pairs. For example, Kato \textit{et al.} \citeyearpar{kato2018compositional} proposed to compose classifiers for unseen verb-noun pairs by leveraging an external knowledge graph and graph convolutional networks.

\noindent{\bf  The Polysemy Problem.} This problem is very common in our daily life. There have been some researches exploring the polysemy problem, e.g. natural language processing \cite{ma2020addressing}, \cite{huang2012improving}, \cite{oomoto2017polysemy} and recommendation systems \cite{liu2019single}. For example, the polysemy problem in  natural language processing  is mainly due to different usages of a word in grammar, e.g., functioned as a verb or a noun \cite{ma2020addressing}.  Besides, each node in a recommendation system could have multiple facets because of the different links with its neighbour nodes, which brings in the node polysemy problem \cite{liu2019single}.

In this paper, we address the verb polysemy problem in HOI detection, i.e., a verb may convey substantially different visual characteristics when associated with different objects. There are also some related areas that may face the same verb polysemy problem, e.g. Action Recognition \cite{simonyan2014two, tran2015learning, damen2018scaling} and Visual Relationship Detection   \cite{lu2016visual, krishna2017visual, Alina2020The, ji2020action}.

Action recognition  methods  aim to recognize human actions from an image or a video. In particular, the EPIC-KITCHENS Dataset \cite{damen2018scaling} is a new large-scale egocentric video benchmark for action recognition tasks.  Besides recognizing the human action from the video, the task on this dataset also involves identifying the interacted object category. It includes 125 verb and 352 noun categories. There are also many verbs suffering from the polysemy problem, e.g. ``hold'', ``open'' and ``close''.

Visual relationship detection  involves detecting and localizing pairs of objects in an image and also classifying the predicate or interaction between each subject-object pair. Different from HOI detection, the subject for each subject-object pair in visual relationship detection can be any object category besides human. Polysemy verbs, e.g. ``carry'', ``ride'', ``hold'' are also common relationships in visual relationship detection. Therefore, visual relationship detection also suffers from the polysemy problem.

Furthermore, as the vast majority of HOI detection methods are based on single images  \cite{Gao2018iCANIA, xu2019learning, li2019transferable, gupta2019no, peyre2019detecting, Qi2018Learning, liao2020ppdm, wang2020learning, ulutan2020vsgnet}, we only consider the verb polysemy problem for image-based HOI detection. However, image-based methods ignore the temporal information, which also provides rich cues to decipher the verb polysemy problem. Therefore, we expect more works on video-based HOI detection in the future.

\noindent{\bf The Exploitation of Language Priors.} Language priors have also been successfully utilized in many computer vision-related fields,
including Scene Graph Generation \cite{lu2016visual, zhang2017visual, gu2019scene, wang2019exploring}, Image Captioning \cite{zhou2019unified, yao2019hierarchy}, and Visual Question Answering \cite{zhou2019unified, gao2019dynamic, marino2019ok}. Moreover, several works \cite{xu2019learning, peyre2019detecting}  have adopted language priors for HOI detection. All of these approaches project visual features and word embeddings to a joint space, which improves HOI detection by exploiting the semantic relationship between similar verbs or HOI categories (e.g. ``drink'' and ``sip'' or ``ride horse'' and ``ride cow''). However, these works do not employ language priors to solve the challenging verb polysemy problem. Compared with the above methods, PD-Net aims to solve the verb polysemy problem by using three novel language prior-based components: Language Prior-guided Channel Attention, Language Prior-based Feature Augmentation, and Polysemy-Aware Modal Fusion.

\begin{figure*}[t]
	\begin{center}
		\includegraphics[width=0.92\textwidth]{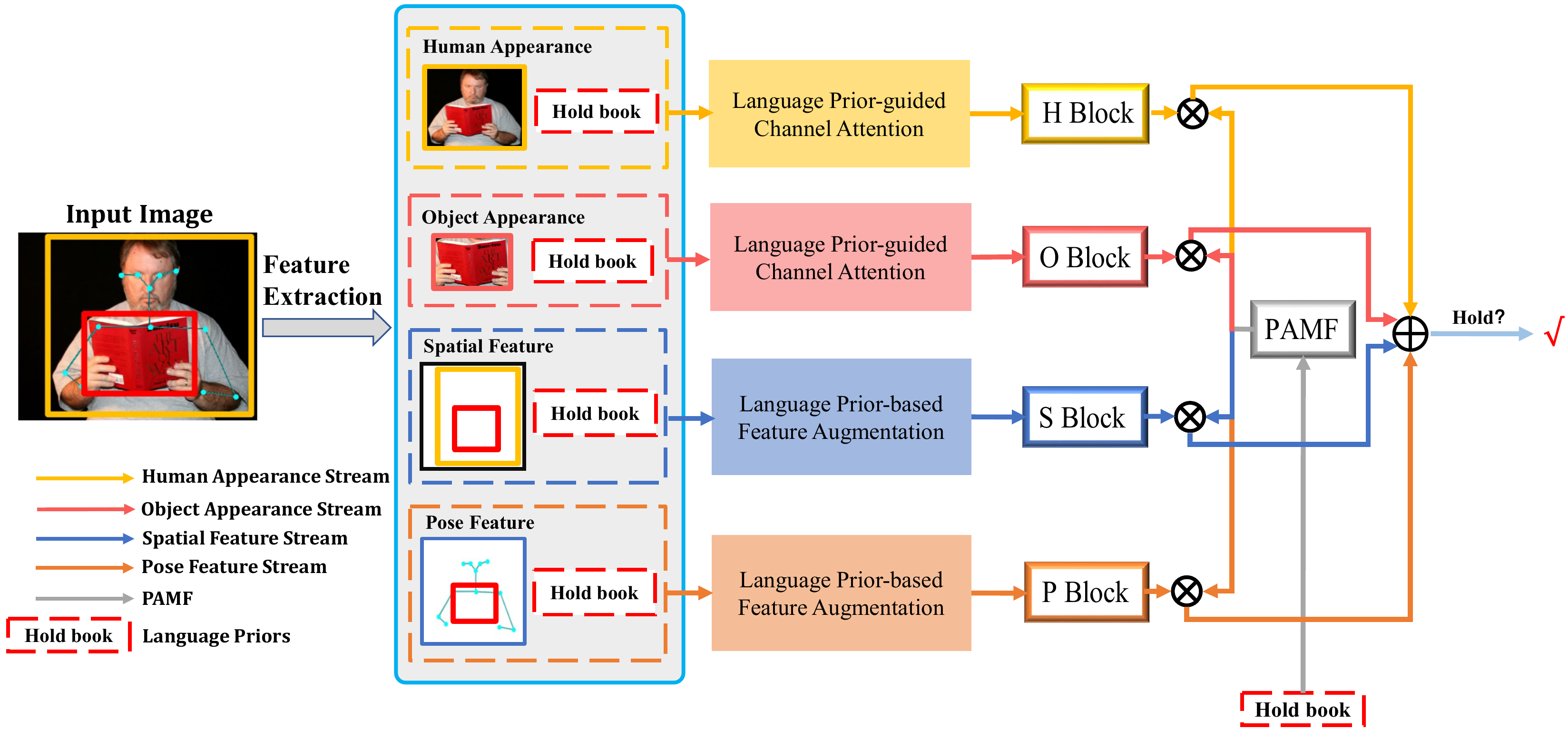}
	\end{center}
	\caption{Overview of the Polysemy Deciphering Network. For the sake of simplicity, only one binary CSP classifier (for ``hold'') is illustrated here. PD-Net takes four feature streams as input: the human appearance stream ({\bf H} stream), the object appearance stream ({\bf O} stream), the human-object spatial stream ({\bf S} stream), and the human pose stream ({\bf P} stream). These four feature streams are first processed by either LPCA or LPFA to be polysemy-aware. They are then sent to the {\bf H}, {\bf O}, {\bf S}, and {\bf P} blocks for binary classification, respectively. Subsequently, the classification scores from the four feature streams are fused using the attention scores produced by ${\bf PAMF}$. Here, $\otimes$ and $\oplus$ denote the element-wise multiplication and addition operations, respectively. }
	\label{Figure:overview}
\end{figure*}
	
\noindent{\bf Attention Models.} Attention mechanisms are becoming a popular component in computer vision tasks, including Image Captioning \cite{chen2017sca, xu2015show, you2016image},
Action Recognition  \cite{girdhar2017attentional, meng2019interpretable}, and Pose Estimation \cite{li2019compression, ye2016spatial}.
Existing studies on attention mechanisms can be roughly divided into three categories: namely, hard regional attention \cite{jaderberg2015spatial, li2018harmonious, wang30dynamic}, soft spatial attention \cite{wang2017residual, pereira2019adaptive, zhu2018attention}, and channel attention \cite{pereira2019adaptive, hu2018squeeze, ding2020multi}.
Hard regional attention methods typically  predict regions of interest (ROIs) first, and then only utilize features in ROIs for subsequent tasks. In comparison, soft spatial attention  and channel attention (CA) models use soft weights to highlight important features in the spatial and channel dimensions, respectively.
There have also been existing works that adopted attention models to assist with HOI detection tasks.
For example, Gao \textit{et al.} \citeyearpar{Gao2018iCANIA} and Wang \textit{et al.} \citeyearpar{wang2019deep} employed an attention mechanism to enhance the human and object features by aggregating contextual information.
Wan  \textit{et al.} \citeyearpar{wan2019pose} proposed the PMFNet model, which adopts human pose and spatial features as cues to infer the importance of each human part.
Ulutan  \textit{et al.} \citeyearpar{ulutan2020vsgnet} used cues derived from a human-object spatial configuration  to highlight the important elements in appearance features.

To the best of our knowledge, few works thus far have made use of the attention mechanism to solve the verb polysemy problem  in HOI detection. Moreover, existing attention models for HOI detection usually employ visual features (e.g. the appearance and pose features) as cues. By contrast, our proposed Language Prior-guided Channel Attention and Polysemy-Aware Modal Fusion adopt language priors as cues; these priors have clear semantic meanings, and are therefore well-suited to resolving the verb polysemy problem.

\section{Method}
\label{sec:method}

We first formulate a basic HOI detection scheme that is adopted by most existing works~\cite{gupta2019no, li2019transferable, wan2019pose, zhou2019relation, li2020detailed}. Then, we explain the verb polysemy problem and formulate the PD-Net framework. Finally, we describe each of the key components in PD-Net.

\subsection{Problem Formulation}

Given an image $I$, human and object proposals are generated using Faster R-CNN~\cite{ren2015faster}.  Each human proposal $h$ and each object proposal $o$ are paired as a candidate for verb classification. HOI detection models ~\cite{gupta2019no, li2019transferable, wan2019pose, zhou2019relation, li2020detailed} then produce a set of verb classification scores for each candidate pair ($h$, $o$). The classification scores for verb $v$ can be represented as follows:

\begin{eqnarray}
    \mathcal{S}_{(h,o,v)} =  \sigma(\sum\limits_{i} {T_{(i,v)}(G_{i}(I,h,o))}) ,
\label{eq:base}
\end{eqnarray}
where $i$ is a subscript that stands for the $i$-th feature type. Function $G_{i}(\cdot)$ denotes the model that produces the $i$-th type of features   ~\cite{gupta2019no, li2019transferable, wan2019pose, zhou2019relation, li2020detailed}. $T_{(i,v)}(\cdot)$  represents the classifier for the verb $v$ using features produced by $G_{i}(\cdot)$. $\sigma(\cdot)$ denotes the sigmoid activation function.

As described in Section \ref{sec:intro}, existing works \cite{gupta2019no, li2019transferable, wan2019pose, zhou2019relation, li2020detailed} suffer from the verb polysemy problem. First, owing to the large intra-class appearance variance for one verb, it is challenging for function $G_{i}(\cdot)$ to learn discriminative visual features for a polysemic verb. Second, the importance of the same feature type may vary dramatically as the objects of interest change. Third, $T_{(i,v)}(\cdot)$ is often shared by different HOI categories with the same verb, which makes it difficult for $T_{(i,v)}(\cdot)$ to capture important visual cues to identify one polysemic verb. Therefore, Eq. (\ref{eq:base}) cannot well address the verb polysemy problem.

Accordingly, in this paper, we propose PD-Net to address the above three problems. Given one human-object pair ($h$, $o$), the classification scores for verb $v$ predicted by PD-Net can be represented as follows:
\begin{eqnarray}
    \mathcal{S}^{\mathbf{PD}}_{(h,o,v)} =  \sigma(\sum\limits_{i} {a_{(i,o,v)}T_{(i,o,v)}(G_{i}(I,h,o, w_{o}, w_{v}))}) ,
\label{eq:pd_net}
\end{eqnarray}
where $w_{o}$ and  $w_{v}$ represent the word embeddings for the object and verb categories, respectively.  $G_{i}(\cdot)$ can leverage word embeddings of one verb-object pair (HOI category) to generate polysemy-aware features. $a_{(i,o,v)}$ denotes attention scores produced by the Polysemy-Aware Modal Fusion module. $T_{(i,o,v)}$ represents the clustering-based object-specific classifiers that are shared by semantically similar HOI categories with the same verb.

In the following, we introduce the framework of PD-Net  based on Eq. (\ref{eq:pd_net}).

\subsection{Overview of PD-Net}

The framework of PD-Net is illustrated in Fig. \ref{Figure:overview}. Similar to existing works  \cite{li2019transferable, gupta2019no,wan2019pose, zhou2019relation, li2020detailed}, four types of visual features are adopted, i.e., human appearance, object appearance, human-object spatial, and human pose features. We construct these four types of features for PD-Net following~\cite{gupta2019no}.
In particular, the human and object appearance features are $K_{A}$-dimensional vectors that are extracted from Faster R-CNN model using human and object bounding boxes, respectively. The human-object spatial feature is a 42-dimensional vector encoded using the bounding box coordinates of one human-object pair. Moreover, we use a pose estimation model~\cite{fang2017rmpe} to obtain the coordinates of 17 keypoints for each human instance. Subsequently, following~\cite{gupta2019no}, the human keypoints and the bounding box coordinates of object proposal are then encoded into a 272-dimensional pose feature vector.

As outlined in Fig. \ref{Figure:singlebinary}, we transform the multi-label verb classification into a set of binary classification problems.  Each of the binary classifiers is used for one verb category verification and only processes features that use this verb as language priors. Moreover, each binary classifier includes a set of {\bf H}, {\bf O}, {\bf S}, and {\bf P} blocks;
Apart from the final layer which is used for verb prediction, the parameters of the other layers in each respective block are shared across different binary classifiers.
Therefore, their overall model size is comparable to an ordinary multi-label classifier. The  binary classifiers mainly differ in terms of their input features and the way in which they combine predictions from the four feature streams.

In the following, we propose four novel components to handle the verb polysemy problem in HOI detection. First, we introduce the Language Prior-guided Channel Attention and Language Prior-based Feature Augmentation modules, which facilitate the four types of features being polysemy-aware. Second, we design the Polysemy-Aware Modal Fusion module that adaptively fuses the prediction scores produced by the four feature streams and obtains the final prediction score for each binary classifier. Finally, we propose a Clustering-based Object-Specific classification scheme that strikes a balance between resolving the verb polysemy problem and reducing the number of binary classifiers in PD-Net.

\subsection{Polysemy-Aware Feature Generation}
\label{sec:LGA}

We here introduce two novel components, i.e. Language Prior-guided Channel Attention and Language Prior-based Feature Augmentation, that generate polysemy-aware features. The two components are denoted as $G_{i}(\cdot)$ in Eq. (\ref{eq:pd_net}).
The language prior we used in this paper is the concatenated word embedding of two words: one word denotes the verb to be identified ($w_{v}$ in Eq. (\ref{eq:pd_net})), while the other one is the detected object in the human-object pair ({$w_{o}$ in Eq. (\ref{eq:pd_net})}).
The word embeddings are generated using the word2vec tool, which was trained on the Google News dataset  ~\cite{mikolov2013distributed}. The dimension of the language prior is 600.

\begin{figure}[t]
\centering
    \includegraphics[width=0.45\textwidth]{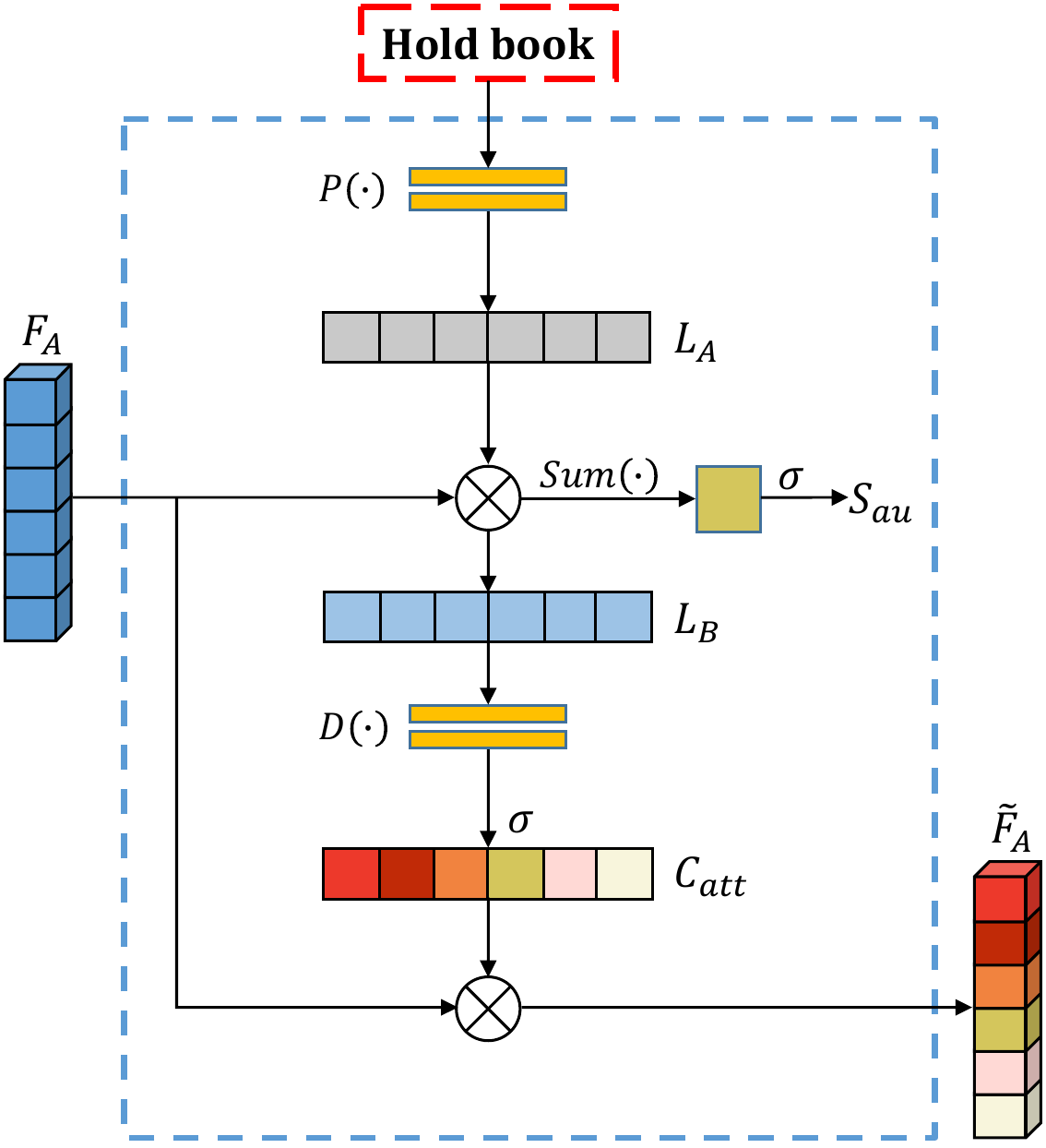}
   \caption{ Model structure of LPCA. $F_{A}$ denotes the human or object appearance feature. $P(\cdot)$ and $D(\cdot)$ are realized by successive fully-connected layers. $\otimes$ denotes the element-wise multiplication operation. }
    \label{Figure:PSFG}
\end{figure}

\noindent{{\bf Language Prior-guided Channel Attention.}} Both the human and object appearance features are usually redundant as only part of their information is involved for a specific HOI category.  One example can be found in Fig. \ref{Figure:diverse_semantic_role}(a) and (b), where the human body parts most relevant to ``play soccer\_ball'' and ``play frisbee'' are the ``feet'' and ``hands'', respectively.  We therefore propose Language Prior-guided Channel Attention (LPCA) to highlight the important elements in human and object appearance features, based on the channel attention scheme and the guidance of language priors.  LPCA is realized through two steps, which are outlined below.

First, we infer the important elements in the human or object appearance feature $F_{A}$ using language priors. $F_{A}$ is extracted from the Faster R-CNN model using a bounding box. As can be seen from Fig. \ref{Figure:PSFG},
the language prior is projected to a $K_{A}$-dimensional vector (denoted as $L_{A}$) via two successive fully-connected layers. The dimension of the first one is set to  $\frac{K_{A}}{2}$; $K_{A}$ is equal to the dimension of $F_{A}$. Similar to \cite{peyre2019detecting},
$L_{A}$ is normalized via its L2 norm. To drive $L_{A}$ to pay attention to important elements in $F_{A}$ regarding to the verb-object pair,
we perform element-wise multiplication between $L_{A}$ and $F_{A}$, as follows:
\begin{eqnarray}
    L_{B} = F_{A}  \otimes  L_{A},
\label{eq:LB}
\end{eqnarray}
and further compute the summation of elements in $L_{B}$:
\begin{eqnarray}
    \mathcal{S}_{au} =  \sigma (Sum(L_{B})),
\label{eq:Sau}
\end{eqnarray}
where $\sigma$ denotes the sigmoid activation function and $Sum(\cdot)$ denotes the summation of all elements in one vector.
During the training stage, we minimize the binary cross-entropy loss between $\mathcal{S}_{au}$ and the binary label for the verb to verify.
During inference, the operation in Eq. (\ref{eq:Sau}) can be ignored. By optimizing the fully-connected layers via the verb verification goal, the value of elements in $L_{A}$ can reflect the importance of the corresponding elements in $F_{A}$.

It is worth noting that the quality of both $L_{A}$ and $L_{B}$ can be affected by the discrepancy between visual features and word embeddings,
since the word embeddings are not specifically designed for computer vision tasks~\cite{xu2019learning}.
Therefore, directly using $L_{B}$ as representation for verb verification may be suboptimal. Consequently, to handle this problem, we propose the following strategy to further enhance the quality of the representations.

Second, we obtain attention scores based on $L_{B}$ via a plain channel attention module:
\begin{eqnarray}
    C_{att} = \sigma (D(L_{B})),
\label{eq:Catt}
\end{eqnarray}
where $ C_{att}$ stands for the final channel attention scores for $F_{A}$. $D(\cdot)$ is realized by two successive fully-connected layers.
By imposing a supervision on $\mathcal{S}_{au}$, elements in $L_{B}$ that are relevant to the verb to verify have large values, which facilitates the learning of an effective channel attention.  For its part, the plain CA module then makes use of the correlation between elements in $L_{B}$ to promote the quality of attention scores. Finally, the polysemy-aware human or object appearance features can be obtained via the following equation:
\begin{eqnarray}
    \tilde{F}_{A} = F_{A} \otimes C_{att},
\label{eq:FA}
\end{eqnarray}

Existing channel attention models, e.g. the SE network \cite{hu2018squeeze}, usually utilize appearance features to generate channel attention scores for the features themselves. Besides, the VSGNet model  \cite{ulutan2020vsgnet}  adopted the human-object spatial features as cues to infer channel attention scores for the appearance feature. Compared with the above two works, LPCA contains clearer semantic information for each HOI. First, the input of LPCA includes language priors, which have clear semantic meaning. Second, LPCA imposes an auxiliary binary cross-entropy loss between $S_{au}$ and the verb to verify. This extra loss enables the visual features for different HOI categories with the same verb to be polysemy-aware according to their specific language priors. Moreover, language priors are adopted in \cite{peyre2019detecting} to construct verb classifiers. In their method, human and object features are fixed.
In comparison, we utilize language priors to generate polysemy-aware features, which are adjustable for each verb-object pair to identify.

Finally, LPCA can also  be regarded as cross-modal attention  \cite{lu2019vilbert} or a conditioning module \cite{perez2018film}.
The differences between LPCA and existing cross-modal attention methods (e.g. ViLBERT \cite{lu2019vilbert}) and conditioning modules (e.g. FiLM \cite{perez2018film}) are illustrated as follows.
First, ViLBERT utilizes a set of image region features and a sequence of word embeddings as input. Then, it conducts cross modal attention based on the standard transformer blocks \cite{vaswani2017attention}, which compute query, key, and value matrices to produce the attention-pooled features. In comparison, LPCA only uses one visual feature vector and one text segment (the concatenation of the word embeddings of the verb and object category) as the input. The Hadamard product between the features $F_{A}$ and $L_{A}$ is utilized to generate the channel attention scores.
Second, FiLM  \cite{perez2018film} carries out a feature-wise affine transformation on a deep model’s intermediate features, directly conditioned on the language prior. In comparison, in order to address the discrepancy between visual features and word embeddings, LPCA generates channel attention scores based on the correlation between visual feature and word embeddings.

\noindent{{\bf Language Prior-based Feature Augmentation.}}
Language Prior-based Feature Augmentation (LPFA) is applied to human-object spatial and human pose features. As illustrated in Fig. \ref{Figure:diverse_semantic_role}(c) and (d), (e) and (f), the spatial and pose features are often vague and vary dramatically for the same verb, meaning that they contain insufficient information. Therefore, we propose LPFA to augment the pose and spatial features. More specifically, we concatenate each of the two features with the 600-dimensional language prior
As a result of this concatenation, the classifiers receive hints that can aid in reducing the intra-class variation of the same verb for the pose and spatial features.

\subsection{Polysemy-Aware Modal Fusion}
\label{sec:PAMF}
As illustrated in Fig. \ref{Figure:overview}, the four feature streams are sent to the {\bf H}, {\bf O}, {\bf S} and {\bf P} blocks, respectively.
The {\bf H} and {\bf O} blocks are constructed using two successive fully-connected layers, while the {\bf S} and {\bf P} blocks are constructed using three successive fully-connected layers.
In the interests of simplicity, the dimension of each hidden fully-connected layer is set as the dimension of its input feature vector.
The output dimension of these four blocks is set to $K_{C}$, which is the number of binary classifiers in PD-Net.

As discussed in Section \ref{sec:intro}, one major challenge posed by the verb polysemy problem is that the relative importance of each of the four feature streams to the identification of the same verb may vary dramatically as the objects change. As shown in Fig. \ref{Figure:diverse_semantic_role}(e) and (f), the human appearance and pose features are most important for detecting $<$$person$ $fly$ $kite$$>$; by contrast, these features are almost invisible and therefore less useful for detecting $<$$person$ $fly$ $airplane$$>$. Therefore, we propose Polysemy-Aware Modal Fusion (PAMF) to generate attention scores that dynamically fuse the predictions of the four feature streams. To the best of our knowledge, PAMF is the first attention module that effectively uses language prior to address the verb polysemy problem in HOI detection.
In more detail, we use the same 600-dimensional word embedding (e.g. ``hold book'') that is implemented in LPCA and LPFA. The language prior is fed into two successive fully-connected layers, the dimensions of which are 48 and 4, respectively. The first fully-connected layer is followed by a ReLU layer, while the second one is followed by a sigmoid activation function.  The output of PAMF is used as attention scores for the four feature streams. In this way, the important feature streams with respect to each HOI category is highlighted, while those that are less important are suppressed. We can see that PAMF is a weight-light module that can be effectively optimized even with limited training data. Moreover, we use the pre-trained word embeddings \cite{mikolov2013distributed} as input for PAMF. These word embeddings have semantical relationships with each other as prior, which further reduces the optimization difficulty of PAMF.

Therefore, Eq. (\ref{eq:pd_net}) can be rewritten as follows:
\begin{eqnarray}
    \mathcal{S}^{\mathbf{PD}}_{(h,o,v)} =  \sigma(\sum\limits_{i \in \{  \mathbf{H}, \mathbf{O}, \mathbf{S}, \mathbf{P}\} } {a_{(i,o,v)}s_{(i,o,v)}}) ,
\label{eq:vs}
\end{eqnarray}
where $i$ denotes one feature stream, while $a_{(i,o,v)}$ is the attention score generated by PAMF for the $i$-th feature stream. $s_{(i,o,v)}$ is the output for verb $v$ generated by the $i$-th feature stream.

\subsection{Clustering-based  Object Specific Verb
Classifiers}
\label{sec:CSP}
Although the above proposed components, i.e. LPCA, LPFA, and PAMF, can help object-shared verb classifiers  to relieve the verb polysemy problem, the defects of object-shared verb classifiers still remains. The essential reason is that HOIs with different objects share the same verb classifier. Under ideal circumstances, object-specific verb classifiers can overcome this problem if sufficient training data exists for each HOI category. However, if we assume that the number of object categories is $|O|$ and the number of verb categories is $|V|$, the total number of their combinations will therefore be $|O|\times|V|$, which is usually very large even if meaningless HOI categories are removed. Therefore, it is too difficult to obtain sufficient train samples for each HOI category. Moreover, due to the class imbalance problem for HOI categories, the object-specific classifiers lack few- and zero-shot learning abilities for HOI categories which have small amount of training data. Therefore, both types of verb classifiers have limitations.

In this subsection, we introduce a novel verb classifier, named Clustering-based object-SPecific (CSP) verb classifiers, which are denoted as $T_{(i,o,v)}$ in Eq. (\ref{eq:pd_net}). CSP classifiers can strike a balance between overcoming the verb polysemy problem and handling  the zero- or few-shot learning problems. The main motivation behind CSP classifiers is that some HOIs tagged with the same verb are both semantically and visually similar, e.g. $<$$person$ $hold$ $sheep$$>$, $<$$person$ $hold$ $horse$$>$, and $<$$person$ $hold$ $cow$$>$; therefore, they can share the same verb classifier, meaning that the number of SP classifiers is consequently reduced. In more detail, we first obtain all meaningful and common HOI categories for each verb, which are available in popular databases such as HICO-DET \cite{chao2018learning} and V-COCO \cite{gupta2015visual}. The number of meaningful HOI categories including the verb $v$ is indicated by ${O}_{v}$. We then use the K-means method \cite{macqueen1967some} to cluster the HOI categories with the same verb ${v}$ into ${C}_{v}$ clusters according to the cosine distance between the word embeddings of the objects. We empirically set the ${C}_{v}$ for each verb as a rounded number of the square root of ${O}_{v}$.

We provide visualization of clustering results for some polysemic verbs in the supplementary file.
During both training and inference, only one CSP classifier is adopted to predict whether the verb is positive for one verb-object pair. The adopted CSP classifier is determined by the object category in the verb-object pair. This clustering strategy is also capable of handling the few- and zero-shot HOI detection problems  \cite{bansal2020detecting}. For example, during testing, a new HOI category $<$$person$ $hold$ $elephant$$>$ can share the same classifier with other HOI categories that have similar semantic meanings (e.g. $<$$person$ $hold$ $horse$$>$).

Besides our automatic way to cluster semantically similar HOI categories, one alternative is to find all possible semantic meanings for a verb from an English dictionary. A dictionary usually elaborates different semantic meanings for each verb. Therefore, the number of clusters for a verb can be determined. Then, one may manually associate each HOI category with one semantic meaning.

One recent work  \cite{bansal2020detecting} also utilized clustering methods to achieve zero-shot HOI detection. There are two differences between the clustering strategies in this work and \cite{bansal2020detecting}. First, we utilize  clustering to build new verb classifiers. In comparison, the clustering strategy in \cite{bansal2020detecting}  is used to generate new training data. Second, we cluster the available HOI categories in one database for each respective verb. In comparison, clustering is conducted only once in  \cite{bansal2020detecting}. Our idea is that some HOI categories are meaningless. By clustering the available HOI categories for each verb, we can obtain more meaningful and fine-grained clustering result.

\subsection{Training and Testing}
\noindent{\bf Training.}
\label{sec:training}
PD-Net can be  conceptualized as a multi-task network. Its loss for the verification of the verb $v$ in one HOI category $(h,v,o)$ can be represented as follows:
\begin{gather}
    \mathcal{L}_{(h,o,v)} =  \mathcal{L}_{BCE}(\mathcal{S}^{\mathbf{PD}}_{(h,o,v)}, {l}_{v}) + \sum\limits_{i \in \{  \mathbf{H}, \mathbf{O}\} } {\mathcal{L}_{BCE}(\mathcal{S}^{i}_{au}},{l}_{v}),
\label{eq:loss}
\end{gather}
 where $\mathcal{L}_{BCE}$ represents binary cross-entropy loss, while ${l}_{v}$ denotes a binary label (${l}_{v} \in $  $\{ 0,1 \}$) for one verb to verify. Moreover, $\mathcal{S}^{\mathbf{H}}_{au}$ and $\mathcal{S}^{\mathbf{O}}_{au}$ denote the output of Eq. (\ref{eq:Sau}) for the human and object appearance features, respectively.

\noindent{\bf Testing.} During testing, we use the same method as that utilized in the training stage to obtain the language priors. Here, the object category in the prior is predicted using Faster
R-CNN (rather than the ground-truth); the verb category in the prior varies for each binary classifier of the verb. Following  \cite{li2019transferable, li2020detailed, ulutan2020vsgnet, wan2019pose}, we also construct an Interactiveness Network (INet) capable of  suppressing pairs without interaction. Finally, the prediction score for one HOI category $(h,v,o)$ is represented as follows:
\begin{equation}
    \mathcal{S}^{\mathbf{HOI}}_{(h,o,v)} =   \mathcal{S}_h \times \mathcal{S}_o \times \mathcal{S}^{\mathbf{PD}}_{(h,o,v)} \times \mathcal{S}^{\mathbf{I}}_{(h,o)},
    \label{eq:final}
\end{equation}
where $\mathcal{S}_h$ and $\mathcal{S}_o$ are the detection scores of human and object proposals, respectively, while $\mathcal{S}^{\mathbf{I}}_{(h,o)}$ denotes the prediction score generated by the pre-trained INet. In the experimental section below, we demonstrate that INet slightly promotes the performance of PD-Net.
\begin{table}[t]
        \centering
        \caption{The number of associated objects and instances for each verb in the HOI-VP dataset.}
        \resizebox{0.26\textwidth}{!}{
        \begin{tabular}{ccc}
        \hline
        Verbs     & \# Objects   &\#  Instances \\
        \hline
        \hline
        carry & 49   &1585 \\
        cross    &8         &611  \\
        fix & 4   &2063 \\
        hold & 229   &13369 \\
        in & 218          &6123 \\
        on & 196   &11752 \\
        make & 8   &147 \\
        open &7   &181 \\
        operate   &4   &46\\
        play & 22 &1119\\
        push & 10 &202\\
        ride & 29 &1734 \\
        swing & 5   &1116 \\
        touch & 18   &154 \\
        use & 29   &3333 \\
        \hline
        \hline
        \end{tabular}}
        \vspace{1mm}
        \vspace{-5mm}
        \label{HOI_P_info}
\end{table}

\section{Experimental Setup}
\label{sec:Experiments_setting}
\subsection{Datasets}
HICO-DET~\cite{chao2018learning} is a large-scale dataset for HOI detection, containing a total of  47,776 images; of these, 38,118 images are assigned to the training set, while the remaining 9,568 images are used as the testing set. There are 117 verb categories, 80 object categories, and 600 common HOI categories overall; moreover, these 600 HOI categories are divided into 138 rare and 462 non-rare categories. Each rare HOI category contains less than 10 training samples.  Each verb is included in an average of five HOI categories.

V-COCO \cite{gupta2015visual} is a subset of MS-COCO \cite{MSCOCO} and contains 2,533, 2,867, and 4,946 images used for training, validation and testing, respectively. There are 24 verb categories and 259 HOI categories in total. Each verb is included in 10 HOI categories on average.

HOI-VerbPolysemy (HOI-VP) is a new database constructed in this paper. To the best of our knowledge, this is the first database to be designed  explicitly for the verb polysemy problem in HOI detection. In more detail, it consists of 15 common verbs (predicates) that have rich and diverse semantic meanings. It also contains 517 common objects in real-world scenarios. Each verb is included in an average of 55 HOI categories, as detailed in Table \ref{HOI_P_info}. In particular, ``in'' and ``on'' are two highly common predicates that are also polysemic in visual relationship detection tasks   \cite{lu2016visual, krishna2017visual, Alina2020The, ji2020action} and are thus both included in the HOI-VP database. There are 21,928 and 7,262 images used for training and testing, respectively. All images  are collected from the VG database \cite{krishna2017visual}, while the corresponding annotations are provided by the HCVRD database \cite{zhuang2017care}. In the HOI-VP dataset, we only use images that were labelled with the 15 predicates listed in Table \ref{HOI_P_info}.  These images and their labels are collected based on the HCVRD dataset. Therefore, the images of HOI-VP can be considered as a subset of HCVRD.

It is worth noting here that the annotations in HCVRD contain noise.  For example, the same verb may be annotated with different words, e.g., ``hold'', ``holds'', and ``holding'', while a similar problem exists for the objects, e.g., ``camera'', ``digital camera'', and ``video camera''.  We therefore merge different annotations for the same verb or object categories, respectively. In the following, we take “hold” as example to explain the correction of annotations. We search for the highly relevant labels with the key word “hold”. Then, we manually check the images with these labels, and make sure that these labels indeed have the same semantic meaning. Finally, we merge the labels of the same semantic meaning with “hold”. The annotation noise for each object category is mainly from its fine-grained attributes. For example, the object ``shirt'' may be labelled as ``black shirt'', ``blue shirt'', and ``stripe shirt''. The merging steps for object labels are the same as those for the verbs.
Some sample images from HOI-VP are illustrated in Fig. \ref{Figure:vishoip}. This database will be made publicly available to expedite research into the verb polysemy problem.

\subsection{Evaluation Metrics}
According to the official protocols \cite{chao2018learning, gupta2015visual}, mean average precision (mAP) is used as the evaluation metric for HOI detection on both HICO-DET and V-COCO datasets. A positive human-object pair must meet the following requirements: first, the predicted HOI category must be the same type as the ground truth; second, both the human and object proposals must have an Intersection over Union (IoU) with the ground truth proposals of more than 0.5. Moreover, there are two mAP modes in HICO-DET, namely the \textbf{Default} (DT) mode and the \textbf{Known-Object} (KO) mode. In the DT mode, we calculate the average precision (AP) for each HOI category in all testing images. In the KO mode, the object categories in all images are known; therefore, we need only to compute the AP for each HOI category from images containing the interested object. For example, we evaluate the AP of $<$$person$ $ride$ $horse$$>$ using only those testing images that contain a ``horse''. Since the images that contain the object category of interest are known, the KO mode is better able to reflect the verb classification ability. For V-COCO, the role mAP \cite{gupta2015visual} ($AP_{role}$) is used for evaluation.

For the HOI-VP database, we use an evaluation protocol similar to that of HICO-DET. As there are as many as 517 object categories in HOI-VP, object detection becomes a challenging task. Accordingly, to reduce the impact of object detection errors, the ground-truth bounding boxes and categories for  both human and object instances are provided. This strategy is similar to the Predicate Classification (PREDCLS) protocol, which has been widely adopted in scene graph generation tasks \cite{zellers2018neural, lin2020gps}.
It facilitates a clean comparison of verb classification ability between different HOI detection models.
\begin{table}[t]
\centering
\caption{Performance comparisons between common object detectors on COCO \cite{MSCOCO}. }
		\resizebox{0.48\textwidth}{!}{
\begin{tabular}{l  c  c }
\hline
Method         & Backbone & AP \\
\hline
Faster R-CNN  \cite{ren2015faster}   & ResNet-152    & 36.7 \cite{chen17implementation}    \\
Faster R-CNN          & ResNet-50-FPN  & 36.8 \cite{massa2018mrcnn} \\
Mask R-CNN \cite{he2017mask} & ResNet-50  &36.9 \cite{Detectron2018}  \\
CenterNet \cite{zhou2019objects}  & Hourglass-104 \cite{newell2016stacked} & 40.3 \cite{zhou2019objects}  \\
Faster R-CNN              &NASNet \cite{zoph2018learning} &43.0 \cite{huang2017speed}\\

\hline
\end{tabular}}
\label{tab:object_detector}
\end{table}
\subsection{Implementation Details}
To facilitate fair comparison with existing works, we consider two popular object detection models for PD-Net. The first of these, Faster R-CNN \cite{ren2015faster} with ResNet-50-FPN \cite{lin2017feature} backbone, attaches a Feature Pyramid Network (FPN) to ResNet-50 \cite{he2016deep} and generates object proposals from the FPN. Based on these proposals, instance appearance features are extracted from the ResNet-50 model. The second model is Faster R-CNN with ResNet-152 backbone \cite{he2016deep}. Here, both instance proposals and appearance features are obtained from the ResNet-152 model. The above two object detectors are trained on the COCO database  \cite{MSCOCO}. As shown in Table \ref{tab:object_detector}, the two object detectors achieve comparable detection performance. Moreover, to facilitate fair comparison with the majority of existing works  \cite{ulutan2020vsgnet, gupta2019no, li2019transferable, peyre2019detecting, Qi2018Learning}, we fix the parameters of both object detectors.
Following~\cite{gupta2019no}, for both human and each object category, we first select the top 10 proposals according to the detection scores after non-maximum suppression. Moreover, the bounding boxes whose detection scores are lower than 0.01 are removed.
The dimension of appearance features for both object detectors, i.e. $K_{A}$, is 2,048.

Utilizing the same approach as existing works \cite{Gao2018iCANIA, xu2019learning, li2019transferable, gupta2019no, peyre2019detecting, Qi2018Learning, liao2020ppdm, wang2020learning, ulutan2020vsgnet, zhong2021glance}, the HOI categories that appear in the training set are set as the meaningful and common HOI categories in each HOI database. The dimension of output layers of the  {\bf H}, {\bf O}, {\bf S},  and {\bf P} blocks, i.e. $K_{C}$, is set to 187, 45, and 83 on the HICO-DET, V-COCO, and HOI-VP databases, respectively; these figures  are equal to the number of CSP classifiers on each respective database. We train PD-Net for 6 (10) epochs using Adam optimizer \cite{kingma2014adam} with a learning rate of 1e-3 (1e-4) on HICO-DET (V-COCO), while on HOI-VP, we train PD-Net for 12 epochs using a learning rate of 1e-3. During testing, we rank the HOI candidate pairs according to their detection scores (obtained via Eq.~(\ref{eq:final})) and calculate mAP for evaluation purposes.

\section{Experimental Results and Discussion}
\label{sec:EXPERIMENTS}
\subsection{Ablation Studies}

\begin{table}[t]
	\centering
\caption{Ablation studies on each component of PD-Net. Full refers to evaluation on all 600 HOI categories in HICO-DET.}
	\resizebox{0.48\textwidth}{!}{
		\begin{tabular}{ccccccccc}
			\hline
			& \multicolumn{6}{c}{Components}  &\multicolumn{2}{c}{Full}  \\
			Methods     & SH&PAMF&LPFA&LPCA&CSP&INet     & DT  & KO\\
			\hline
			Our Baseline    &\checkmark&-&-&-&-&-                     &17.57 &23.07 \\
			\hline
			\multirow{4}{*}{Incremental}
			&\checkmark&\checkmark&-&-&-&-                      &18.86   &24.43  \\
            &\checkmark&\checkmark&\checkmark&-&-&-                      &19.38    &24.64  \\
			&\checkmark&\checkmark&\checkmark&\checkmark&-&-                 &20.71 & 24.85  \\
            &-&\checkmark&\checkmark&\checkmark&\checkmark&-                 &21.77 &\textbf{26.98}  \\
			\hline
			\multirow{4}{*}{Drop-one-out}	&-&-&\checkmark&\checkmark&\checkmark&-         &20.14 &25.65\\
			&-&\checkmark&-&\checkmark&\checkmark&-      &21.37 &26.15 \\
			&-&\checkmark&\checkmark&-&\checkmark&-             &19.32 &24.30  \\
            &\checkmark&\checkmark&\checkmark&\checkmark&-&-                 &20.71 & 24.85  \\
			\hline
			\vspace{0.mm}
			PD-Net &-&\checkmark&\checkmark&\checkmark&\checkmark&\checkmark  & \textbf{22.37} &26.86  \\
			\hline
	\end{tabular}}
	\label{tab:ablation1}
\end{table}

To demonstrate the effectiveness of each proposed component in PD-Net, we perform ablation studies on the HICO-DET database. In Table \ref{tab:ablation1}, the baseline is constructed by removing Language Prior-guided Channel Attention (LPCA), Language Prior-based Feature Augmentation (LPFA), and Polysemy-Aware Modal Fusion (PAMF) from PD-Net; we also replace Clustering-based object-SPecific (CSP) classifiers with object-SHared (SH) classifiers. The other settings for the baseline remain the same as in PD-Net. For both models, the Faster R-CNN with ResNet-152 backbone is used for object detection. Experimental results are summarized in Table \ref{tab:ablation1}. From these results, we can make the following observations.

\noindent{\bf Effectiveness of PAMF.} Polysemy-Aware Modal Fusion is designed to decipher the verb polysemy by assigning larger weights to more important feature types for each HOI category. As shown in Table~\ref{tab:ablation1}, PAMF promotes the performance of the baseline  by 1.29\% and 1.36\% mAP in DT and KO modes, respectively.

\noindent{\bf Effectiveness of LPFA.} Language Prior-based Feature Augmentation is used to provide hints for the classifier in order to reduce the intra-class variation of the pose and spatial features by augmenting them with language priors. When LPFA is incorporated, HOI detection performance is promoted by  0.52\% and 0.21\%  mAP in DT and KO modes, respectively.

\noindent{\bf Effectiveness of LPCA.} The appearance features are redundant for HOI detection. Language Prior-guided Channel Attention  is proposed to generate polysemy-aware appearance features. As can be seen from Table \ref{tab:ablation1}, LPCA promotes the HOI detection performance by a clear margin of 1.33\% and 0.21\% in DT and KO modes, respectively.

\noindent{\bf Effectiveness of CSP Classifiers.} Clustering-based object-SPecific  classifiers can relieve the verb polysemy problem by assigning the same verb classifier to semantically similar HOI categories. As shown in Table \ref{tab:ablation1}, CSP classifiers  improve the HOI detection performance by  1.06\% and 2.13\% mAP in DT and KO modes, respectively.

\noindent{\bf Drop-one-out Study.} We further perform a drop-one-out study in which each proposed component is removed individually. These experimental results further demonstrate that each component is indeed helpful to promote HOI detection performance.

Finally, when INet is integrated, the mAP of PD-Net in the DT mode is further promoted by 0.60\%. However, the mAP in the KO mode does not improve. This is because INet can assist PD-Net by suppressing candidate pairs without interactions, which are usually caused by incorrect or redundant object proposals in the DT mode.
However, the KO mode is comparatively less affected by object detection errors; therefore, PD-Net can achieve high performance without the assistance of INet in this mode. This experiment demonstrates that the strong performance of PD-Net is primarily a result of its excellent verb classification ability.


\begin{table}[t]
\centering
\caption{Comparisons  with  one variant  of  the  language  prior. }
\resizebox{0.28\textwidth}{!}{
\begin{tabular}{l  c c c}
\hline
Language Prior  & DT (Full)  &KO (Full) \\
\hline
Verb Only         &19.98 &24.99 \\
Verb  + Object  & \textbf{22.37} &\textbf{26.86}\\
\hline
\end{tabular}}
\label{tab:ablation_lp}
\end{table}

\subsection{Comparisons with Variants of PD-Net}
\subsubsection{Comparisons with Variants of the Language Prior}
In this experiment, we remove the word embedding of the object category from the language prior so that only the word embedding of the verb category to identify is used as input for PAMF, LPFA, and LPCA. As shown in Table \ref{tab:ablation_lp}, without the word embedding of the object category, the performance of PD-Net drops by a large margin of  2.39\% (1.87\%) mAP in DT (KO) mode. These experimental results indicate that the word embedding of the object category in the language prior is an important hint to decipher the verb polysemy problem.

\subsubsection{Comparisons with Variants of LPCA}
In this experiment, we compare the performance of Language Prior-guided Channel Attention (LPCA)  with five possible variants: namely, Plain Channel Attention (CA), `w/o $\mathcal{S}_{au}$', `w/o $\mathcal{C}_{att}$', `$D([L_{A}, F_{A}])$' and FiLM \cite{perez2018film}. The other implementation details of PD-Net are kept the same for different variants. Plain CA means that we  feed the appearance feature $F_{A}$ directly into a plain CA module, i.e. $D(\cdot)$ in Fig. \ref{Figure:PSFG}, and obtain  $\tilde{F}_{A}$. `w/o $S_{au}$'  involves removing the extra supervision signal $S_{au}$ from LPCA, while `w/o ${C}_{att}$' means that we directly use $L_{B}$ in Fig. \ref{Figure:PSFG} as the input of the $\mathbf{H}$ and $\mathbf{O}$ blocks in Fig. \ref{Figure:overview}, without the further processing by the  plain CA module. `$D([L_{A}, F_{A}])$' means that we use the concatenation of $L_{A}$ and $F_{A}$ as the input for function $D(\cdot)$ to generate channel attention scores in Eq. (\ref{eq:Catt}).  FiLM means that we replace LPCA with a FiLM layer \cite{perez2018film}.
Experimental results are tabulated in Table \ref{tab:ablation2}. In this table, `w/o LPCA' is a baseline that removes the entire LPCA module from PD-Net. From these results, we can make the following observations.

First, the plain CA module alone slightly promotes the performance of PD-Net. One main reason for this is that the plain CA module has very little ability to identify important elements in the appearance features for each HOI category.
\begin{table}[t]
\centering
\caption{Performance comparisons  with  variants for LPCA. $[\cdot]$ represents feature concatenation operation.}
\resizebox{0.32\textwidth}{!}{
\begin{tabular}{l  c c c}
\hline
Methods                & DT (Full)  &KO (Full) \\
\hline
w/o LPCA               &19.82 &23.94  \\
w/o $\mathcal{S}_{au}$ &19.41 &23.32 \\
Plain CA               &19.97 &24.11\\
$D([L_{A}, F_{A}])$               &20.62 &24.45\\
FiLM \cite{perez2018film} &20.90 &25.29 \\
w/o $C_{att}$ & 21.99  &26.34 \\
LPCA  & \textbf{22.37} &\textbf{26.86}\\

\hline
\end{tabular}}
\label{tab:ablation2}
\end{table}
Second, without  supervision from $S_{au}$, the performance of LPCA degrades dramatically. Compared to the plain CA module, this setting adopts language priors to provide cues regarding the  channel-wise importance of $F_{A}$ for each HOI category. However, it receives only implicit supervision from the binary score $\mathcal{S}^{\mathbf{PD}}_{(h,o,v)}$ in Eq. (\ref{eq:final}), which is too weak to optimize LPCA's parameters. We therefore observe degraded performance after the extra supervision $\mathcal{S}_{au}$ is removed.

Third, `w/o $C_{att}$' obtains better performance than both the `Plain CA' and  `w/o $S_{au}$' settings. However, its performance is still lower than that of our proposed LPCA by 0.38\% and 0.52\% mAP in DT and KO modes, respectively. This may be because $L_{A}$ is obtained via projection from the language prior.
As word embeddings are not specifically designed for computer vision tasks, $L_{A}$ may not always be reliable and the quality of $L_{B}$ is affected~\cite{xu2019learning}. Therefore, further processing $L_{B}$ using the plain CA module is helpful.

Fourth, LPCA outperforms $D([L_{A}, F_{A}])$ by significant margins in both DT and KO modes. There are two main reasons for this. First, the concatenation operation significantly increases the model size of the channel attention module, which makes the model more difficult to train. Second, with the optimization on $\mathcal{S}_{au}$, $L_{B}$ can provide more direct hints about important channels in  $F_{A}$ to the verb to verify than $[L_{A}, F_{A}]$.

Fifth, LPCA also outperforms FiLM \cite{perez2018film} by significant margins in both DT and KO modes.This is because the feature-wise affine transformation in \cite{perez2018film} is directly conditioned on the language prior; therefore, it can be affected by the semantic misalignments between visual features and word embeddings. In comparison, LPCA can better address the discrepancy because it produces channel attention scores conditioned on the correlation between language priors and visual features. Moreover, FiLM  \cite{perez2018film} is only supervised by the final classification loss of the model while LPCA is also optimized by an auxiliary supervision on $S_{au}$.

In comparison, our proposed LPCA achieves the best performance for the following reasons. First, it adopts language priors to provide hints regarding the channel-wise importance of $F_{A}$ for each HOI category. Second, it imposes direct supervision to the attention module, which helps to more effectively optimize the model parameters. Third, it refines the attention vector obtained from the  language priors using a plain CA module, which enhances the quality of the channel attention vectors. The above experimental results and analysis demonstrate the effectiveness of LPCA.
\begin{table}[t]
\centering
\caption{Performance comparisons between SH, SP, and CSP verb classifiers.}
		\resizebox{0.48\textwidth}{!}{
\begin{tabular}{l  c  c  c  c  c  c}
\hline
         & \multicolumn{3}{c}{DT Mode} & \multicolumn{3}{c}{KO Mode} \\
Methods         & Full & Rare & Non-Rare  & Full & Rare & Non-Rare\\
\hline
SH & 21.06      &16.45  &22.43 & 24.83      &19.89  &26.30 \\
SP  &20.91 & 15.03 &22.66 & 24.67      &17.83  &26.71 \\
CSP & \textbf{22.37} & \textbf{17.61} & \textbf{23.79} & \textbf{26.86} & \textbf{21.70} & \textbf{28.44}\\
\hline
\end{tabular}}
\label{tab:ablation3}
\end{table}
\subsubsection{Comparisons with Variants of Verb Classifiers}  To further demonstrate the advantages of clustering-based object-specific (CSP) classifiers, we compare their performance with that of object-SHared (SH) and object-SPecific (SP) verb classifiers. To facilitate fair comparison,  other settings of PD-Net remain unchanged. Experimental results are tabulated in Table \ref{tab:ablation3}.
\begin{table*}[t]
\centering
\caption{Performance Comparisons on HICO-DET. The best performance on different backbones are bolded in different colours.}
\resizebox{0.98\textwidth}{!}{
\begin{tabular}{l   c  c  c  c  c  c  c  }
\hline
                &             & \multicolumn{3}{c}{DT Mode}  &\multicolumn{3}{c}{KO Mode} \\
Methods   & Object Detector Backbone   & Full & Rare & Non-Rare  & Full & Rare & Non-Rare \\
\hline
\hline

Shen \MakeLowercase{\textit{et al.}} \cite{Shen2018Scaling} & VGG-19 & 6.46  & 4.24  & 7.12   & -   & -   & -\\
InteractNet \cite{Gkioxari2017Detecting} & ResNet-50-FPN & 9.94  & 7.16     & 10.77     & -   & -   & -\\
GPNN \cite{xu2019learning, Qi2018Learning}  & ResNet-152  & 13.11  & 9.34     & 14.23     & -   & -   & -\\
iHOI \cite{8848601} & ResNet-50-FPN &13.39  &9.51  &14.55 & -   & -   & -\\
Xu \MakeLowercase{\textit{et al.}}  \cite{xu2019learning} & ResNet-50-FPN &14.70  &13.26  &15.13 & -   & -   & -\\
iCAN \cite{Gao2018iCANIA}     & ResNet-50-FPN & 14.84  & 10.45     & 16.15     & 16.26   & 11.33   & 17.73\\
Wang \MakeLowercase{\textit{et al.}} \cite{wang2019deep}  & ResNet-50-FPN  & 16.24  & 11.16     & 17.75     & 17.73   & 12.78 & 19.21\\
No-Frills \cite{gupta2019no} & ResNet-152    & 17.18  & 12.17     & 18.68     & -   & -   & -\\
TIN \cite{li2019transferable}  & ResNet-50-FPN      & 17.22 & 13.51   & 18.32 & 19.38 & 15.38 & 20.57 \\
RPNN \cite{zhou2019relation} & ResNet-50 (Mask R-CNN)  &17.35 &12.78 &18.71  & -   & -   & -\\
PMFNet \cite{wan2019pose}     & ResNet-50-FPN & 17.46  & 15.65     & 18.00     & 20.34   & 17.47   &21.20\\
Peyre \MakeLowercase{\textit{et al.}}  \cite{peyre2019detecting} & ResNet-50-FPN &19.40 &14.60 &20.90  & -   & -   & -\\
IP-Net  \cite{wang2020learning} & ResNet-50-FPN  &19.56 &12.79 &21.58  & 22.05   & 15.77   & 23.92\\
VSGNet \cite{ulutan2020vsgnet} & NASNet \cite{zoph2018learning} &19.80 &16.05 &20.91  & -   & -   & -\\
2D-RN ($\mathcal{S}^{2D}$) \cite{li2020detailed} & ResNet-50-FPN &19.98 &\textbf{16.97} &20.88 &22.56 &19.48 &23.48 \\
PPDM \cite{liao2020ppdm} & Hourglass-104 &21.73 &13.78 & 24.10  & 24.58   & 16.65   & 26.84\\
\hline
\hline
Our baseline (SH)  & ResNet-50-FPN &17.27  &12.27  &18.77  &23.07  &18.29  &24.50  \\
$\mathbf{PD}$-$\mathbf{Net}$      & ResNet-50-FPN  & \textbf{20.76}  & 15.68  & \textbf{22.28} & \textbf{25.59} & \textbf{19.93} & \textbf{27.28} \\
Our baseline (SH) & ResNet-152 &17.57  &12.67  &19.04  &23.07  &17.45  &24.75  \\
$\mathbf{PD}$-$\mathbf{Net}$      & ResNet-152  & {\textcolor[RGB]{255,0,0}{\textbf{22.37}}} & {\textcolor[RGB]{255,0,0}{\textbf{17.61}}} & {\textcolor[RGB]{255,0,0}{\textbf{23.79}}} & {\textcolor[RGB]{255,0,0}{\textbf{26.86}}} & {\textcolor[RGB]{255,0,0}{\textbf{21.70}}} & {\textcolor[RGB]{255,0,0}{\textbf{28.44}}}\\
\hline
\end{tabular}}
\label{tab:hico}
\vspace{-0.5mm}
\end{table*}
It is shown that SH classifiers outperform SP classifiers by 1.42\% (2.06\%) mAP in DT (KO) mode for rare HOI categories.  This is because SH classifiers enable these rare HOI categories to share verb classifiers with other HOI categories that have sufficient training data. By comparison, SP classifiers are better able to relieve the verb polysemy problem for the HOI categories that have sufficient training data. Therefore, the SP classifiers outperform SH classifiers by 0.23\% (0.41\%) mAP in DT (KO) mode for non-rare HOI categories.

In comparison, CSP classifiers achieve superior performance on both rare and non-rare HOI categories. This is due to the same verb classifiers being assigned to semantically similar HOI categories, enabling HOI categories with few training samples to share verb classifiers with those HOI categories that have sufficient training data. Moreover, different verb classifiers are adopted for semantically different HOI categories, which is helpful to overcome the verb polysemy problem. Overall, CSP classifiers outperform SH and SP classifiers by 1.31\% (2.03\%) and 1.46\% (2.19\%) mAP in DT (KO) mode for the full HOI categories, respectively.
The superior performance on rare HOI categories demonstrates the effectiveness of CSP classifiers in the few-shot learning ability. We also further justify the effectiveness of CSP classifiers in terms of zero-shot HOI detection in Section B of the supplementary file.

\subsection{Comparisons with State-of-the-Art Methods}

We compare the performance of PD-Net with state-of-the-art methods on three databases, namely HICO-DET, V-COCO, and HOI-VP. Experimental results are summarized in  Table \ref{tab:hico}, Table \ref{VCOCO}, and Table \ref{tab:hoip}, respectively.

\subsubsection {Performance comparisons on HICO-DET} As shown in Table \ref{tab:hico}, PD-Net outperforms state-of-the-art methods by significant margins using both object detector backbones. It is worth noting that one most recent method, i.e. PPDM, adopts CenterNet with Hourglass-104 backbone \cite{zhou2019objects} as the object detector. As shown in Table \ref{tab:object_detector}, this object detector significantly outperforms the two Faster R-CNN object detectors utilized in our model.  To facilitate fair comparison, we mainly compare PPDM with PD-Net in the KO mode, as this mode is less affected by object detection results. As shown in Table \ref{tab:hico}, PD-Net outperforms  PPDM  in KO mode by significant margins of 2.28\%, 5.05\% and 1.60\% mAP on the full, rare and non-rare HOI categories, respectively. Moreover, PD-Net also outperforms PPDM by 0.64\% in the DT mode on the full HOI categories.

Moreover, as shown in Table \ref{tab:hico} and Table \ref{tab:object_detector}, the object detector adopted by another recent  work \cite{ulutan2020vsgnet} is also much stronger than ours. But PD-Net still outperforms this model by large margins of 2.57\% (22.37\%-19.80\%), 1.56\% (17.61\%-16.05\%), and 2.88\%
(23.79\%-20.91\%)  mAP in the DT mode on the full, rare, and non-rare HOI categories, respectively.

\begin{figure}[t]
\centering
    \includegraphics[width=0.42\textwidth]{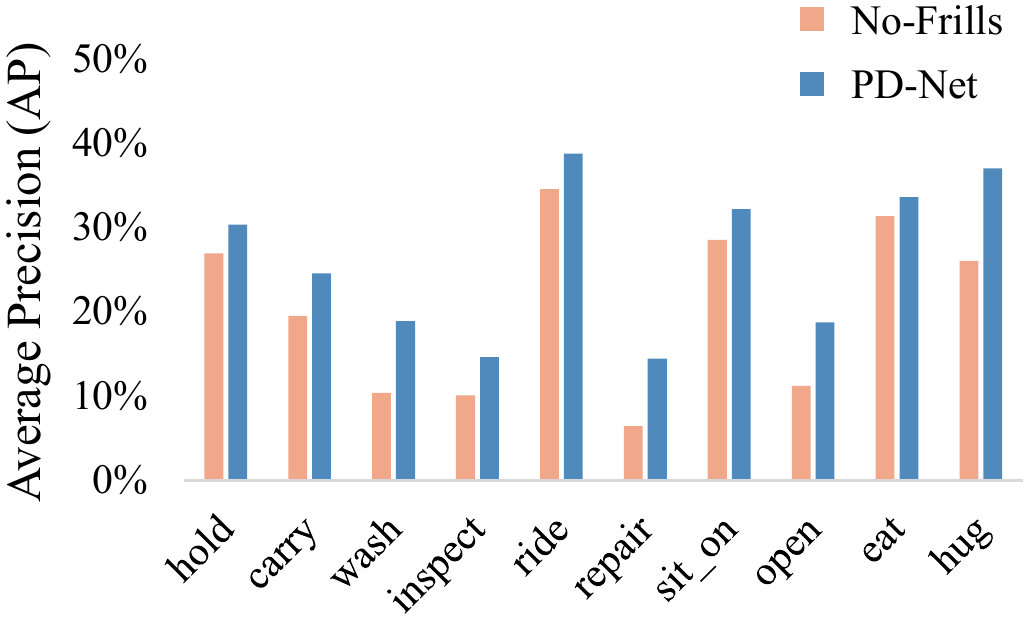}
   \caption{The top 10 verbs that are  most likely to suffer from the polysemy problem on HICO-DET. }
    \label{Figure:HICO_top10}
\end{figure}

Finally, with a similar multi-stream representation network and object detector backbone (ResNet-50-FPN), PD-Net outperforms one very recent model 2D-RN \cite{li2020detailed} by 3.03\%  (25.59\%-22.56\%) and 0.78\% (20.76\%-19.98\%) in mAP on the full HOI categories in the KO and DT modes, respectively.
Another advantage of PD-Net compared with 2D-RN is that PD-Net requires no extra human annotation.
Besides, 3D human pose and 3D object locations are also utilized to improve 2D-RN during inference in \cite{li2020detailed}. To facilitate fair comparisons, we only compare the performance of PD-Net with methods that utilize 2D human pose and 2D object locations during inference.

To further illustrate the advantage of PD-Net in deciphering the verb polysemy problem,  we present  the top 10 verbs (from the total 117 verbs in HICO-DET) ranked by the number of HOI categories in which each verb is included in Fig. \ref{Figure:HICO_top10}. The largest number of HOI categories associated with the same verb (``hold'') is 61. As these verbs are more likely to be affected by the visual polysemy problem, we therefore compare the performance of PD-Net with one state-of-the-art method \cite{gupta2019no} on these verbs.
This method is chosen as it is very similar to our baseline. Results show that  PD-Net achieves superior performance on all of these top 10 verbs.
\begin{table}[t]
        \centering
        \caption{ Performance comparisons on V-COCO \cite{gupta2015visual}. $^{\circ}$  denotes methods that we reproduce. ${\dagger}$ denotes methods that adopt human part features as input.}
        \resizebox{0.48\textwidth}{!}{
        \begin{tabular}{ccc}
        \hline
        Methods    & Object Detector Backbone   & $AP_{role}$ \\
        \hline
        \hline
        Gupta \MakeLowercase{\textit{et al.}}  \cite{gupta2015visual, Gkioxari2017Detecting}  & ResNet-50-FPN         &31.8 \\
        InteractNet \cite{Gkioxari2017Detecting} & ResNet-50-FPN   &40.0 \\
        GPNN \cite{xu2019learning, Qi2018Learning}      & ResNet-152            &44.0  \\
        iCAN   \cite{Gao2018iCANIA}   & ResNet-50-FPN    &45.3\\
        iHOI  \cite{8848601} & ResNet-50-FPN & 45.8 \\
        Xu  \MakeLowercase{\textit{et al.}}  \cite{xu2019learning}  & ResNet-50-FPN &45.9\\
        No-Frills$^{\circ}$  \cite{gupta2019no}   & ResNet-152  &46.7\\
        Wang  \MakeLowercase{\textit{et al.}} \cite{wang2019deep} & ResNet-50-FPN &47.3 \\
        RPNN$^{\dagger}$  \cite{zhou2019relation} & ResNet-50 (Mask R-CNN) &47.5 \\
        TIN  (RP$_{D}$C$_{D}$) \cite{li2019transferable} &ResNet-50-FPN &47.8\\
        C-HOI \cite{zhou2020cascaded} & ResNet-50 & 48.3\\
        IP-Net  \cite{wang2020learning} & ResNet-50-FPN &51.0\\
        VSGNet \cite{ulutan2020vsgnet}  & NASNet \cite{zoph2018learning} &51.7\\
        PMFNet$^{\dagger}$ \cite{wan2019pose} & ResNet-50-FPN & 52.0\\

        \hline
        \hline
        Our baseline (SH)  & ResNet-50-FPN                      &48.5\\
        $\mathbf{PD}$-$\mathbf{Net}$      & ResNet-50-FPN  &\textbf{52.3} \\
        $\mathbf{PD}$-$\mathbf{Net}^{\dagger}$     & ResNet-50-FPN &\textbf{53.3}\\
        Our baseline (SH)   & ResNet-152                        &48.2\\
        $\mathbf{PD}$-$\mathbf{Net}$       & ResNet-152 &{\textcolor[RGB]{255,0,0}{\textbf{52.2}}}\\
        \hline
        \end{tabular}}
        \vspace{-5mm}
        \label{VCOCO}
\end{table}

\begin{table}[t]
        \centering
        \caption{Per verb class AP (\%) comparisons between PMFNet$^{\dagger}$ and PD-Net$^{\dagger}$  on V-COCO. ${\dagger}$ denotes methods that adopt human part features as input.}
        \resizebox{0.36\textwidth}{!}{
        \begin{tabular}{ccc}
        \hline
        Verbs     & PMFNet$^{\dagger}$ \cite{wan2019pose}   &PD-Net$^{\dagger}$ \\
        \hline
        \hline

        hold-obj & 44.01          &\textbf{45.07} \\
        sit-instr & 29.51   &\textbf{31.86} \\
        ride-instr     &70.33         &\textbf{71.80}  \\
        look-obj   &45.22    &\textbf{46.72}\\
        hit-instr & 76.30 &\textbf{78.57} \\
        hit-obj &\textbf{52.28} &50.28\\
        eat-obj  & 44.55 &\textbf{47.41} \\
        eat-instr  & 5.93 &\textbf{6.94} \\
        jump-instr &\textbf{53.39} &52.75 \\
        lay-instr & 26.40 &\textbf{28.25}\\
        talk on phone-instr & 54.69 &\textbf{56.64} \\
        carry-obj & 44.24 &\textbf{45.64}\\
        throw-obj &\textbf{49.76} & 47.82\\
        catch-obj & 54.11 &\textbf{55.01}\\
        cut-instr & 40.08 &\textbf{42.69}\\
        cut-obj &\textbf{40.01} & 39.24\\
        work on computer-instr &67.39 &\textbf{67.98}\\
        ski-instr &\textbf{53.04} &52.59\\
        surf-instr & 80.47 &\textbf{80.95}\\
        skateboard-instr & 86.81 &\textbf{88.00}\\
        drink-instr & 46.76 &\textbf{53.84}\\
        kick-obj & 72.70 &\textbf{74.50}\\
        read-obj & 36.80 &\textbf{39.07}\\
        snowboard-instr & 74.33 &\textbf{76.55} \\
        \hline
        \hline
        mean & 52.05 &\textbf{53.34} \\
        \hline
        \end{tabular}}
        \vspace{1mm}
        \vspace{-5mm}
        \label{VCOCO_cmp}
\end{table}

\subsubsection{Performance comparisons on V-COCO}
To boost the performance on V-COCO, we add another appearance feature stream to both our baseline and PD-Net, following \cite{wan2019pose}. There are consequently a total of five feature streams for experiments on V-COCO.  This new stream extracts appearance features from union boxes composed of human-object pairs. We further apply LPCA to this feature stream in PD-Net. As shown in Table \ref{VCOCO}, PD-Net outperforms state-of-the-art methods by clear margins with both object detectors. In particular, PD-Net outperforms one of the most recently developed methods, i.e. VSGNet~\cite{ulutan2020vsgnet}. As shown in Table~\ref{tab:object_detector},  the object detector utilized by VSGNet is much stronger  \cite{huang2017speed} than ours; nevertheless, PD-Net still outperforms VSGNet by clear margins, as indicated in Table~\ref{VCOCO}.

Moreover, PD-Net outperforms another particularly strong model, named PMFNet \cite{wan2019pose} by 0.3\% (52.3\%-52.0\%) in mAP. The excellent performance of PMFNet may benefit from the use of human part features. Therefore, we adopt the same five feature streams that include the human part features in PMFNet as input for PD-Net; this model is denoted as PD-Net$^{\dag}$ in Table  \ref{VCOCO}. The contributions in this paper remain unchanged. PD-Net$^{\dag}$ outperforms PMFNet by a large margin of 1.3\% (53.3\%-52.0\%) in mAP. Moreover, as shown in Table \ref{VCOCO_cmp}, we compare the performance between PD-Net$^{\dagger}$ and PMFNet on each of the 24 verbs in V-COCO. Here, our method demonstrates superior performance on the vast majority of verb classes.

\begin{table}[t]
\centering
\caption{Performance Comparisons on HOI-VP. }
\resizebox{0.48\textwidth}{!}{
\begin{tabular}{l   c  c    }
\hline
Method   & Feature Extraction Backbone   & mAP \\
\hline
\hline
iCAN \cite{Gao2018iCANIA}     & ResNet-50 & 58.32 \\
TIN \cite{li2019transferable} & ResNet-50      & 60.66\\
No-Frills \cite{gupta2019no} & ResNet-152   & 61.05 \\
Peyre \MakeLowercase{\textit{et al.}}   \cite{peyre2019detecting} & ResNet-50-FPN &61.46 \\
PMFNet \cite{wan2019pose}     & ResNet-50-FPN & 62.30   \\

\hline
\hline
Our baseline (SH)   &ResNet-50 &61.18  \\
$\mathbf{PD}$-$\mathbf{Net}$      &ResNet-50 & \textbf{63.11}\\
Our baseline (SH) &ResNet-50-FPN &61.10  \\
$\mathbf{PD}$-$\mathbf{Net}$      &ResNet-50-FPN & {\textcolor[RGB]{0,0,255}{\textbf{63.66}}}\\
Our baseline (SH)   & ResNet-152 &60.69 \\
$\mathbf{PD}$-$\mathbf{Net}$      & ResNet-152  & {\textcolor[RGB]{255,0,0}{\textbf{63.60}}}\\
\hline
\end{tabular}}
\label{tab:hoip}
\vspace{-0.5mm}
\end{table}

\subsubsection {Performance comparisons on HOI-VP} We next compare the performance of PD-Net with some recent open-source methods, i.e. iCAN  \cite{Gao2018iCANIA}, TIN \cite{li2019transferable}, No-Frills \cite{gupta2019no}, and PMFNet \cite{wan2019pose} on the new HOI-VP database. We also reproduce the method presented in \cite{peyre2019detecting}  that achieves high performance on the HICO-DET database. To facilitate fair comparison, we compare the performance of PD-Net with each of these methods using the same feature extraction backbone, respectively. As shown in Table \ref{tab:hoip}, PD-Net  consistently achieves the best performance out of all compared methods. In particular,  PD-Net outperforms one recent powerful model PMFNet by a clear margin of 1.36\% (63.66\%-62.30\%) in mAP.  As the verbs (predicates) in the HOI-VP database are very common and polysemic in real world scenarios, experimental results on this database demonstrate the superiority of PD-Net to overcome the verb polysemy problem.

\subsection{Qualitative Visualization Results}

\begin{figure}[t]
\centering
    \includegraphics[width=0.48\textwidth]{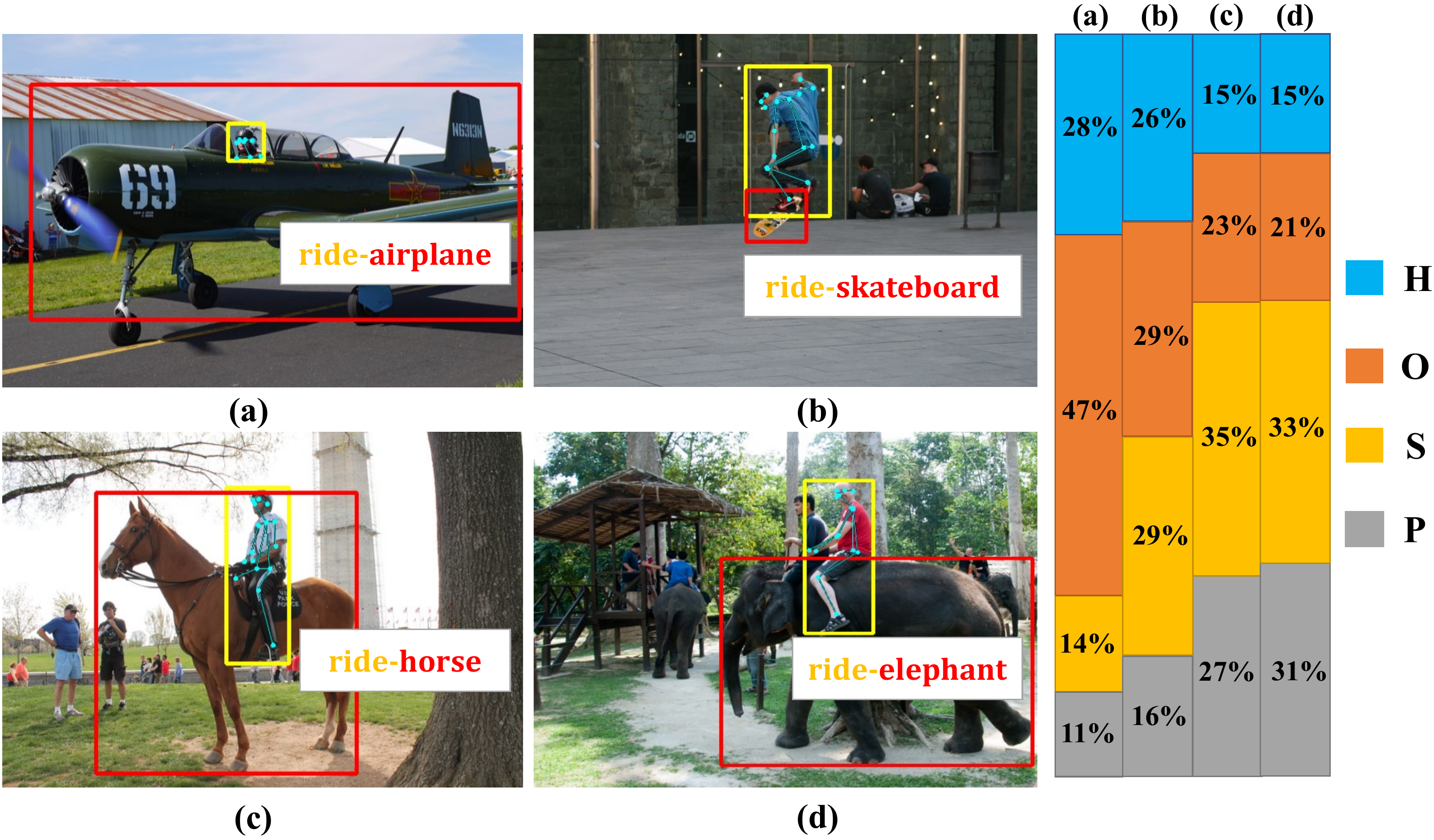}
   \caption{Attention scores produced by PAMF on four types of features. HOI categories in this figure share the same verb, i.e. ``ride''. \textbf{H}, \textbf{O}, \textbf{S}, and \textbf{P} denote human appearance, object appearance, spatial feature, and human pose feature respectively.}
    \label{Figure:PADM_results}
\end{figure}

Fig. \ref{Figure:PADM_results} illustrates attention scores produced by PAMF for four types of features. HOI categories in this figure share the verb ``ride'', but differ dramatically in semantic meanings. The ``person'' proposal in Fig. \ref{Figure:PADM_results}(a) is very small and severely occluded while the ``airplane'' proposal is very large; therefore, object appearance feature is much more important for verb classification than the human appearance feature. In Fig. \ref{Figure:PADM_results}(b), both the spatial feature and the object appearance feature play important roles in determining the verb. Attention scores for Fig. \ref{Figure:PADM_results}(c) and (d) are similar, as $<$$person$ $ride$ $horse$$>$ and $<$$person$ $ride$ $elephant$$>$ are indeed close in semantics.

\begin{figure*}[t]
	\begin{center}
		\includegraphics[width=1.0\textwidth]{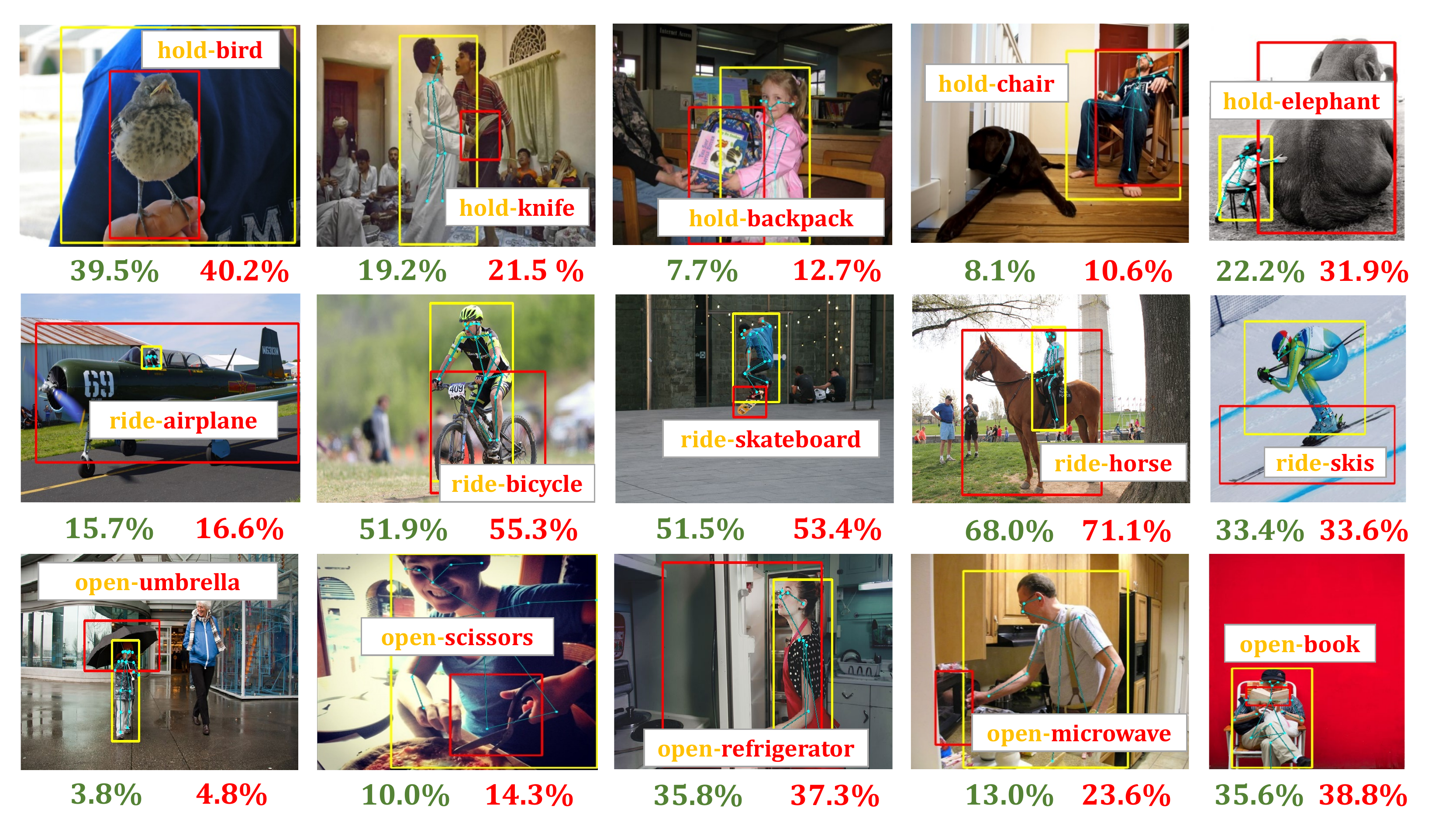}
	\end{center}
    \caption{Visualization of PD-Net's advantage in deciphering the verb polysemy problem on HICO-DET. We randomly select three verbs affected by the polysemy problem: ``hold'' (top row), ``ride'' (middle row), and ``open'' (bottom row).  The {\textcolor[RGB]{0,139,69}{green}} and  {\textcolor[RGB]{255,0,0}{red}} numbers denote the AP of our baseline and PD-Net respectively for the same HOI category. }
	\label{Figure:vishico}
\end{figure*}

\begin{figure*}[t]
	\begin{center}
		\includegraphics[width=1.0\textwidth]{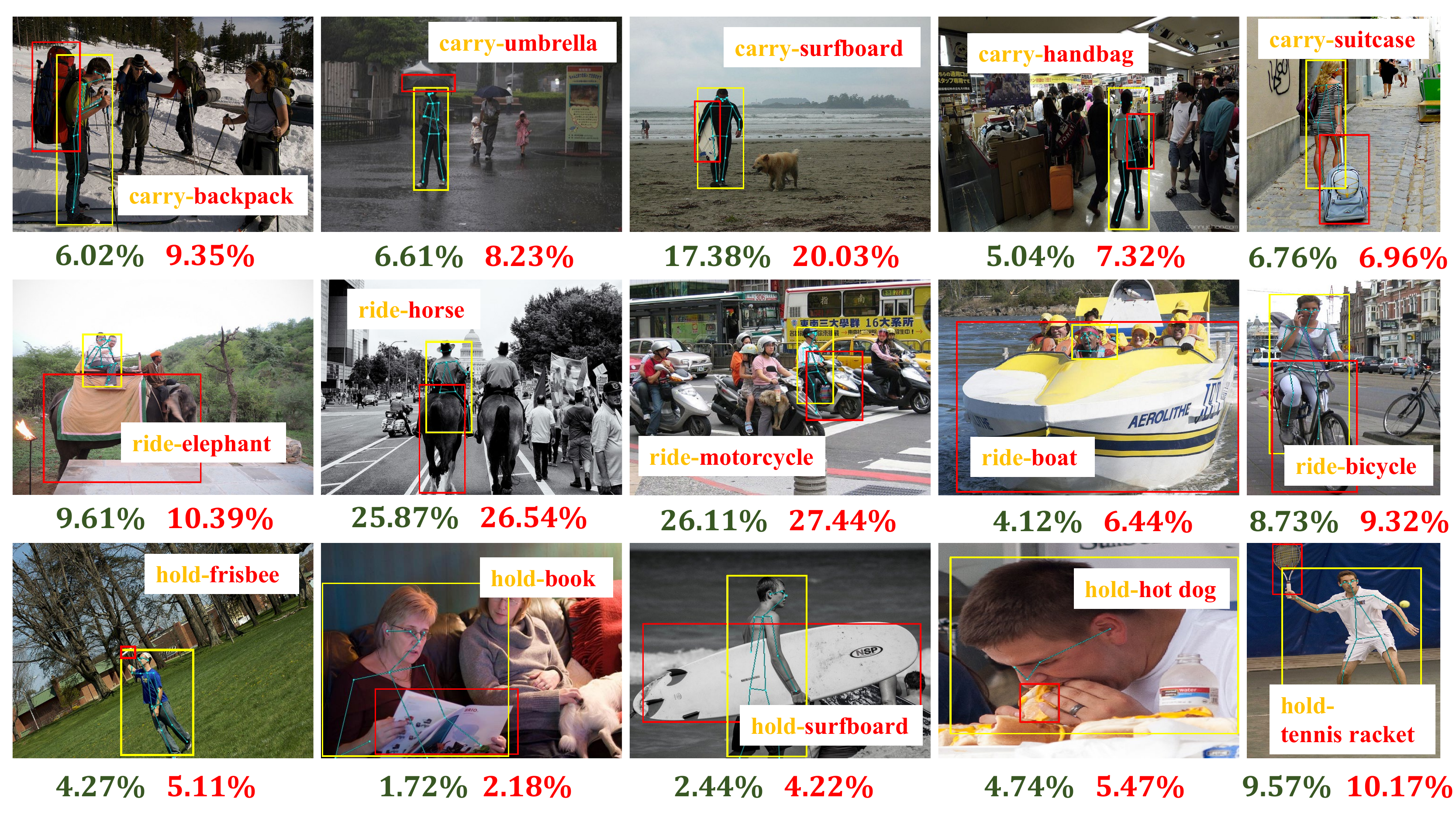}
\end{center}
\caption{Visualization of PD-Net's advantage in deciphering the verb polysemy problem on V-COCO. We randomly select three verbs affected by the polysemy problem: ``carry'' (top row), ``ride'' (middle row), and ``hold'' (bottom row).}
	\label{Figure:visvcoco}
\end{figure*}

\begin{figure*}[t]
	\begin{center}
		\includegraphics[width=1.0\textwidth]{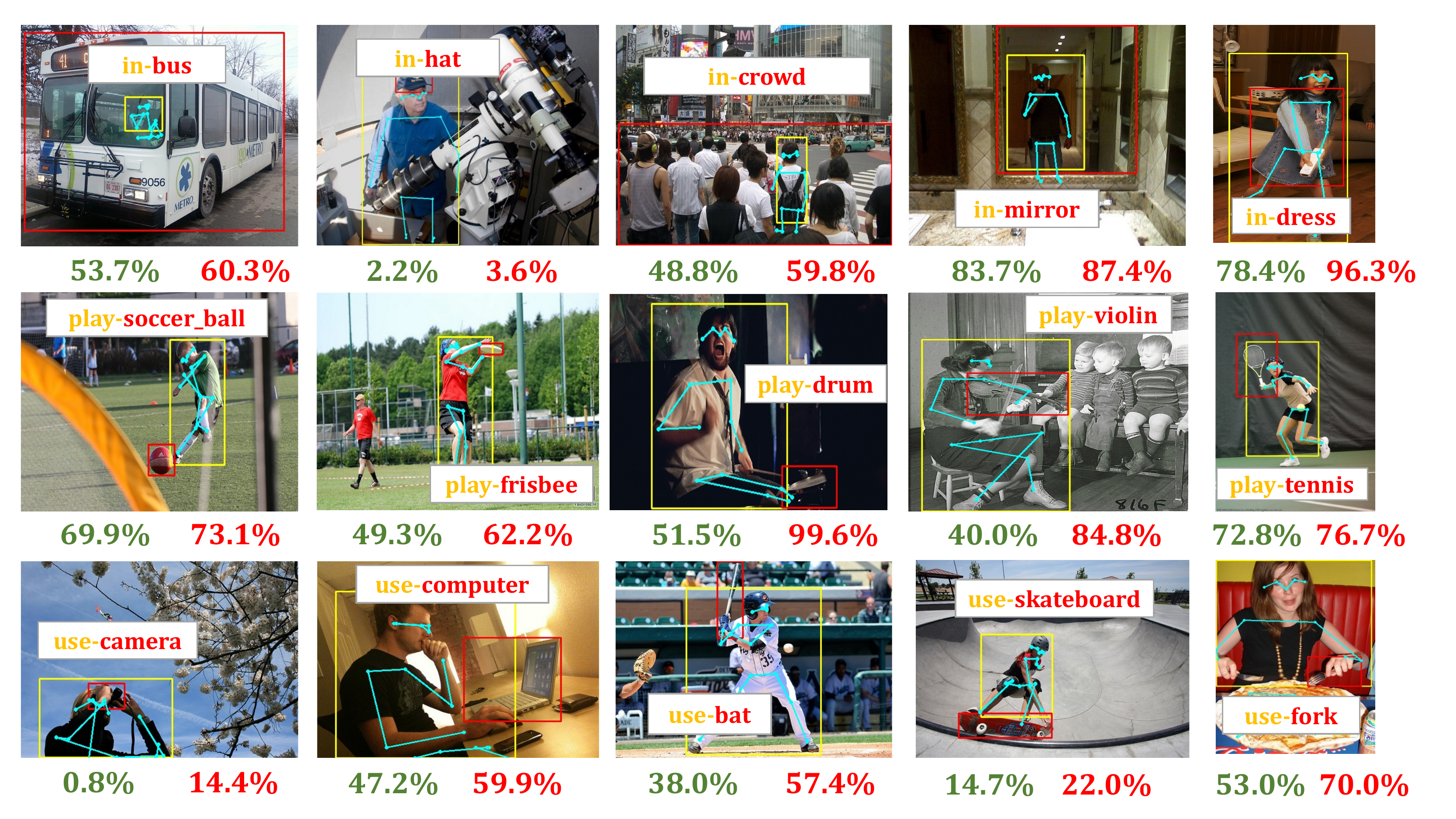}
    \end{center}
\caption{Visualization of PD-Net's advantage in deciphering the verb polysemy problem on HOI-VP. We randomly select three verbs affected by the polysemy problem: ``in'' (top row), ``play'' (middle row), and  ``use'' (bottom row).}
	\label{Figure:vishoip}
\end{figure*}

Fig. \ref{Figure:vishico}, Fig. \ref{Figure:visvcoco},  and  Fig. \ref{Figure:vishoip} provide more examples that demonstrate PD-Net's advantages in deciphering the verb polysemy problem on  HICO-DET, V-COCO, and HOI-VP, respectively. The performance gain by PD-Net compared with our baseline reaches 10.6\%, 3.33\%,  and 48.1\% in AP for the ``open microwave'', ``carry backpack'', and ``play drum'' category on the three datasets, respectively.

\section{Conclusion}
\label{sec:conclusion}
The verb polysemy problem is relatively underexplored and is sometimes even ignored in existing works for HOI detection. Accordingly, in this paper, we propose a novel model named PD-Net, which significantly mitigates the challenging verb polysemy problem. PD-Net includes four novel components: Language Prior-guided Channel Attention, Language Prior-based Feature Augmentation, Polysemy-Aware Modal Fusion, and Clustering-based Object Specific classifiers. Language Prior-guided Channel Attention and Language Prior-based Feature Augmentation are introduced to generate polysemy-aware visual features. Polysemy-Aware Modal Fusion highlights important feature types for each HOI category. The Clustering-based Object Specific classifiers not only relieve the verb polysemy problem, but also is capable of handling the zero- or few-shot learning problems. Exhaustive ablation studies are performed to demonstrate the effectiveness of these components. We further develop and present a new dataset, named HOI-VP, that is specifically designed to expedite the research on the verb polysemy problem for HOI detection.  Finally, by decoding the verb polysemy, we achieve state-of-the-art methods on the three HOI detection benchmarks. In the future, we will study the verb polysemy problem in related tasks to HOI detection, e.g., visual relationship detection and action recognition.

\noindent{\bf Acknowledgement}
This work was supported by the National Natural Science Foundation of China under Grant 62076101, 61702193, and U1801262, the Program for Guangdong Introducing Innovative and Entrepreneurial Teams under Grant 2017ZT07X183, the Natural Science Fund of Guangdong Province under Grant 2018A030313869, the Science and Technology Program of Guangzhou under Grant 201804010272, the Guangzhou Key Laboratory of Body Data Science under Grant 201605030011, and the Fundamental Research Funds for the Central Universities of China under Grant 2019JQ01.

%
%




\clearpage

\begin{appendices}
\section*{Appendix}

This supplementary material includes four sections.
Section \ref{sec:cluster_vis} visualizes  the clustering results for several polysemic verbs and shows the statistics for polysemic verbs in the HICO-DET database.
Section \ref{sec:zero_shot} shows the experimental results of Clustering-based object-SPecific (CSP) verb classifiers in terms of zero-shot HOI detection.
Section \ref{sec:PAMF_vis_more} provides more randomly selected samples to show the attention scores predicted by Polysemy-Aware Modal Fusion (PAMF).
Section \ref{sec:cmp_cls} compares three types of verb classifiers, i.e. object-shared (SH), object-specific (SP), and CSP verb classifiers, based on our baseline model.

\section{Visualization of Clustering  Results }
\label{sec:cluster_vis}

\begin{figure*}[t]
	\begin{center}
		\includegraphics[width=0.96\textwidth]{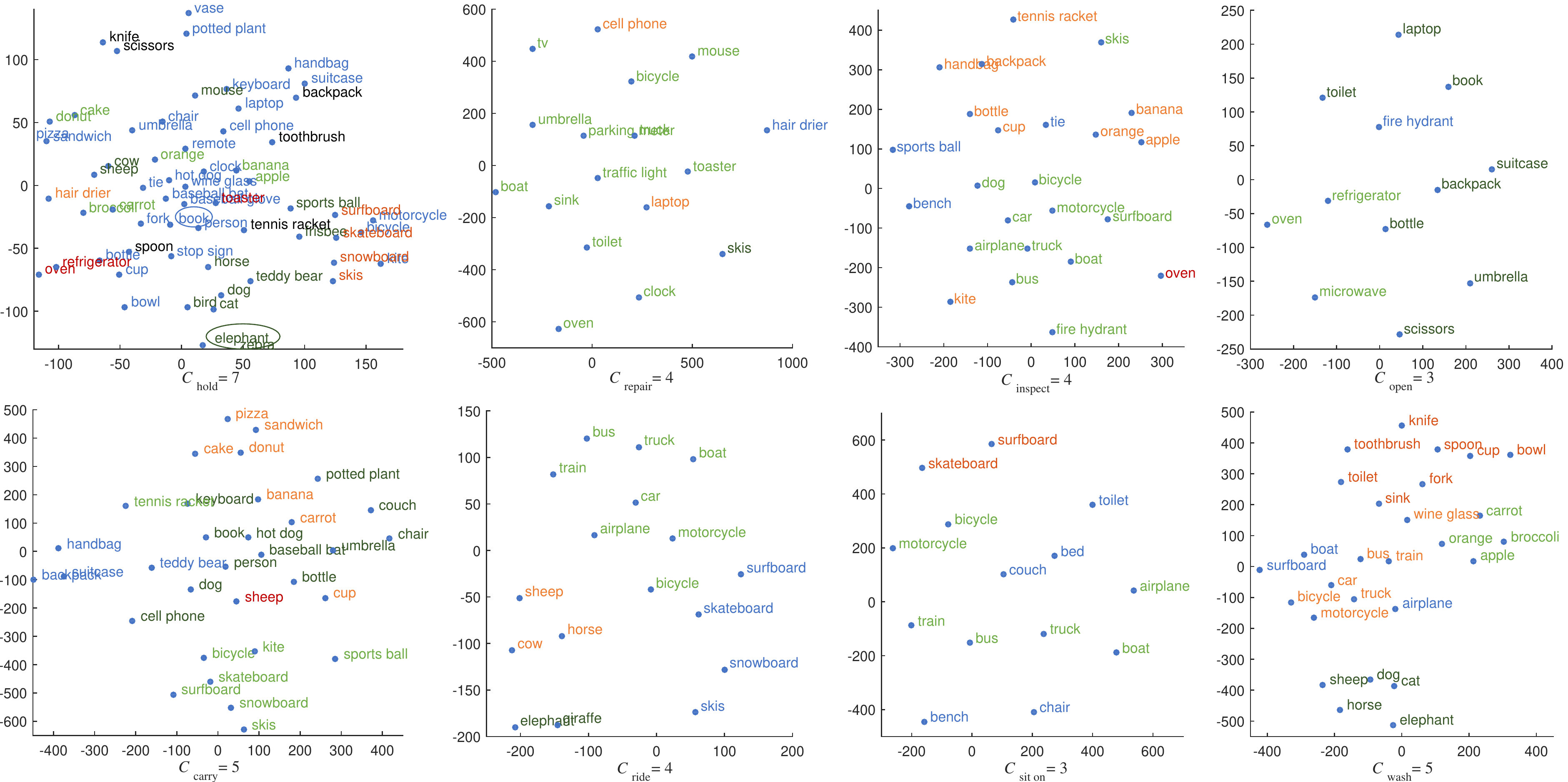}
	\end{center}
	\caption{Visualization of clustering results for polysemic verbs via t-SNE \cite{maaten2008visualizing}. $C_{v}$ denotes the number of clusters for verb $v$.  Objects in the same cluster are tagged with the same colour. }
	\label{Figure:cluster_vis}
\end{figure*}

We visualize the clustering results for multiple polysemic verbs in Fig. \ref{Figure:cluster_vis}.

We observe that the more object categories one verb associates, the more polysemic it presents visually. Therefore, we propose to cluster semantically similar HOI categories for each verb and set the cluster number as the rounded square root of the number of associated objects to a verb. It is also clear that objects in the same cluster are usually of similar functions or properties, which demonstrates the effectiveness of our clustering strategy in terms of distinguishing different  semantic  meanings for a verb.

For example, as illustrated in Fig. 1(c) and (d) of the main paper,  ``hold book'' and ``hold elephant'' present different visual characteristics for the verb ``hold'',  we can see that ``book'' and ``elephant'' (highlighted in circle) are in different clusters in the visualization results for the verb ``hold''.

In Table \ref{tab:1}, we show the statistics for polysemic verbs in the HICO-DET database \cite{chao2018learning}. According to our clustering strategy, there are 16 verbs that contain over 3 semantic meanings and involves nearly 58.5\% HOI categories.

\begin{table}[h]
\centering
\caption{Statistics of Polysemic Verbs on HICO-DET. }
\resizebox{0.48\textwidth}{!}{
\begin{tabular}{l  c c c}
\hline
\# Associated Object Catgories  &\# Verb (Ratio)  &\# HOI category (Ratio) \\
\hline
$\geq$9         &16 (13.7\%) &351 (58.5\%) \\
$\geq$4    & 38 (32.5\%) &471 (78.5\%) \\
$\geq$2 & 78 (66.7\%) &561 (93.5\%) \\
\hline
\end{tabular}}
\label{tab:1}
\end{table}

\section{Performance Comparisons on Zero-Shot HOI Detection}
\label{sec:zero_shot}

In this section, we compare the performance of PD-Net with previous works  \cite{Shen2018Scaling}, \cite{bansal2020detecting} in both seen-object and unseen-object settings that were introduced in \cite{bansal2020detecting} . The experimental results for the two settings are summarized in Table \ref{tab:zero_shot_seen_obj} and Table \ref{tab:zero_shot_unseen_obj}, respectively.

In the seen-object setting, we follow \cite{bansal2020detecting} to select 5 random sets of 120 unseen HOI categories from the total 600 HOI categories, and ensure that each object involved in these 120 HOI categories occurs at least one time in the remaining 480 ones. During training, we only use samples of the 480 seen HOI categories to train our model. We report the mean performance over these 5 random sets of 120 unseen HOI categories in Table \ref{tab:zero_shot_seen_obj}. Moreover, \cite{bansal2020detecting} only utilizes human appearance and human-object spatial feature as input. To facilitate fair comparisons, we also only use these two types of features for PD-Net. We can observe that PD-Net outperforms \cite{bansal2020detecting} by significant margins, which verifies its effectiveness for zero-shot HOI detection.

In the unseen-object setting, as \cite{bansal2020detecting} did not provide the selected unseen objects, we randomly select 12 objects from the total 80 object classes as unseen objects. Accordingly, there are 100 HOI categories associated with these 12 objects, which are set as unseen HOIs. Following \cite{bansal2020detecting}, we only use the samples of the remaining 500 HOI categories to train our model.
As there are no training samples for the unseen objects, during inference, we adopt the same generic object detector as that in \cite{bansal2020detecting} to generate object proposals for the unseen objects. The experimental results are provided in Table \ref{tab:zero_shot_unseen_obj}. We can see that PD-Net significantly outperforms \cite{bansal2020detecting} in terms of zero-shot HOI detection. We also replace CSP classifiers with object-shared (SH) classifiers in PD-Net, which is denoted as PD-Net (SH) in Table D. SH classifiers share training samples for HOI categories with the same verb; therefore, it has zero-shot learning ability for unseen HOI categories. PD-Net achieves better performance than PD-Net (SH) for unseen HOI categories, which further demonstrates the effectiveness of CSP classifiers in terms of zero-shot HOI detection.

\begin{table}[t]
\centering
\caption{Performance comparisons on zero-shot HOI detection in the seen-object setting. Full means all categories including both Seen and Unseen objects.}
\resizebox{0.4\textwidth}{!}{
\begin{tabular}{l  c c c c}
\hline
Method &Unseen  &Seen &Full\\
\hline
Shen \MakeLowercase{\textit{et al.}} \cite{Shen2018Scaling}          &5.62 &- &6.26 \\
Bansal \MakeLowercase{\textit{et al.}}  \cite{bansal2020detecting}      & 11.31 	&12.74	&12.45 \\
PD-Net (SH)  & 15.28	&17.41	&16.98 \\
PD-Net & 15.95 	&18.80	&18.23 \\
\hline
\end{tabular}}
\label{tab:zero_shot_seen_obj}
\end{table}

\label{sec:zero_shot}
\begin{table}[t]
\centering
\caption{Performance comparisons on zero-shot HOI detection in the unseen-object setting. Full means all categories including both Seen and Unseen objects.}
\resizebox{0.4\textwidth}{!}{
\begin{tabular}{l  c c c c}
\hline
Method &Unseen  &Seen &Full\\
\hline
Bansal \MakeLowercase{\textit{et al.}}  \cite{bansal2020detecting}      & 11.22 	&14.36	&13.84 \\
PD-Net (SH)  & 14.85	&16.89	&16.55\\
PD-Net & 15.49 	&17.07	&16.80 \\
\hline
\end{tabular}}
\label{tab:zero_shot_unseen_obj}
\end{table}

\begin{table}[t]
\centering
\caption{Performance comparisons between three types of verb classifiers. Full refers to
evaluation on all 600 HOI categories in HICO-DET.}
\resizebox{0.28\textwidth}{!}{
\begin{tabular}{l  c c c}
\hline
Methods  &DT (Full)  &KO (Full) \\
\hline
SH &17.57 &23.07\\
SP    & 17.51 &22.77\\
CSP & 18.30 &23.75 \\
SH + INet &18.43 &23.04\\
SP + INet   & 17.92 &22.65\\
CSP + INet & 19.02 &23.60 \\
\hline
\end{tabular}}
\label{tab:baseline}
\end{table}

\section{Qualitative Visualization Results }
\label{sec:PAMF_vis_more}
\begin{figure*}[t]
	\begin{center}
		\includegraphics[width=0.96\textwidth]{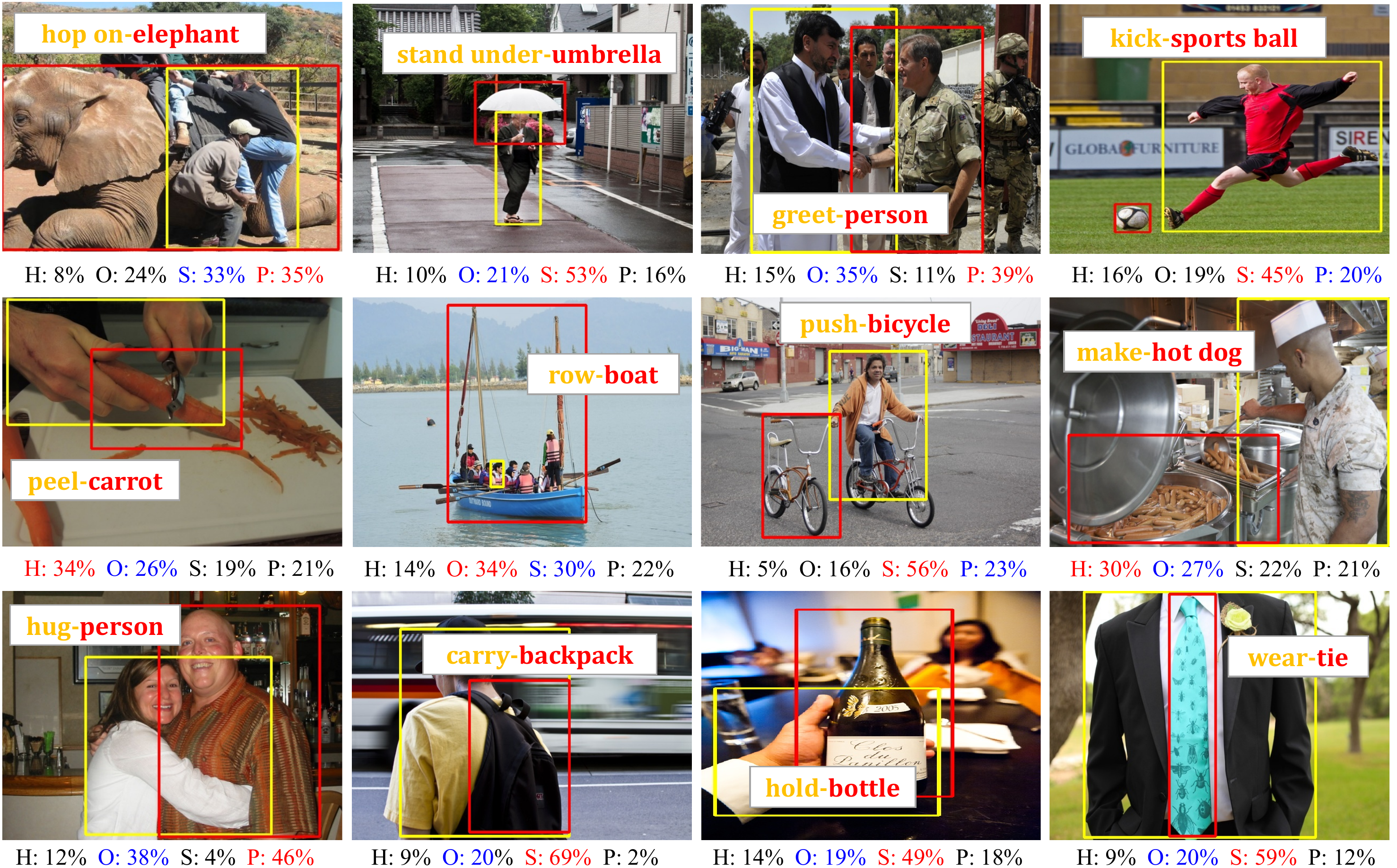}
	\end{center}
	\caption{Visualizations of attention scores produced by PAMF on four types of features.  H, O, S, and P denote human appearance, object appearance, spatial feature, and human pose feature, respectively. The top 1 and top 2 important feature types and their attention scores for one HOI category are highlighted using red and blue, respectively. }
	\label{Figure:PAMF_vis}
\end{figure*}

In Fig. \ref{Figure:PAMF_vis}, we provide more randomly selected samples to show the attention scores predicted by Polysemy-Aware Modal Fusion (PAMF).

\section{Performance Comparisons between Three Verb Classifiers}
\label{sec:cmp_cls}
In this section, we compare the performance of the three verb classifiers, i.e. object-shared (SH), object-specific (SP), and clustering-based object-specific (CSP) verb classifiers. In particular, we replace the SH classifiers in our baseline with SP classifiers and CSP classifiers, respectively. The experiment results are summarized in Table \ref{tab:baseline}.  We can see that our proposed CSP classifier consistently outperforms the other two types of classifiers, which further demonstrate its effectiveness.


%
%



\end{appendices}
\end{document}